\documentclass[sigconf]{acmart}

\copyrightyear{2025}
\acmYear{2025}
\setcopyright{acmlicensed}\acmConference[MM '25]{Proceedings of the 33rd ACM International Conference on Multimedia}{October 27--31, 2025}{Dublin, Ireland}
\acmBooktitle{Proceedings of the 33rd ACM International Conference on Multimedia (MM '25), October 27--31, 2025, Dublin, Ireland}
\acmDOI{10.1145/3746027.3755452}
\acmISBN{979-8-4007-2035-2/2025/10}

\usepackage{multirow}
\usepackage{changepage}
\usepackage[skip=2pt]{caption}
\usepackage{utfsym}
\usepackage{colortbl}

\AtBeginDocument{%
  }

\acmConference[MM '25]{Make sure to enter the correct
  conference title from your rights confirmation email}{October 27--31,
  2025}{Dublin, Ireland}

\acmSubmissionID{3846}

\begin{document}

\title{LIDAR: Lightweight Adaptive Cue-Aware Fusion Vision Mamba for Multimodal Segmentation of Structural Cracks}


\author{Hui Liu}
\affiliation{%
  \institution{Tianjin University of Technology}
  \city{Tianjin}
  \country{China}
}
\email{liuhui1109@stud.tjut.edu.cn}
\orcid{0009-0009-5742-0202}

\author{Chen Jia}
\authornote{Chen Jia and Xu Cheng are the co-corresponding authors.}
\affiliation{%
  \institution{Tianjin University of Technology}
  \city{Tianjin}
  \country{China}
}
\email{jiachen@email.tjut.edu.cn}
\orcid{0000-0002-7535-479X}

\author{Fan Shi}
\affiliation{%
  \institution{Tianjin University of Technology}
  \city{Tianjin}
  \country{China}
}
\email{shifan@email.tjut.edu.cn}
\orcid{0000-0003-2074-0228}

\author{Xu Cheng}
\authornotemark[1]
\affiliation{%
  \institution{Tianjin University of Technology}
  \city{Tianjin}
  \country{China}
}
\email{xu.cheng@ieee.org}
\orcid{0000-0002-4724-5748}

\author{Mengfei Shi}
\affiliation{%
  \institution{Tianjin University of Technology}
  \city{Tianjin}
  \country{China}
}
\email{smf@stud.tjut.edu.cn}
\orcid{0009-0006-3761-1931}

\author{Xia Xie}
\affiliation{%
  \institution{Hainan University}
  \city{Haikou}
  \country{China}
}
\email{shelicy@hainanu.edu.cn}
\orcid{0009-0002-6890-3663}

\author{Shengyong Chen}
\affiliation{%
  \institution{Tianjin University of Technology}
  \city{Tianjin}
  \country{China}
}
\email{sy@ieee.org}
\orcid{0000-0002-6705-3831}


\renewcommand{\shortauthors}{Hui Liu et al.}

\begin{abstract}
Achieving pixel-level segmentation with low computational cost using multimodal data remains a key challenge in crack segmentation tasks. Existing methods lack the capability for adaptive perception and efficient interactive fusion of cross-modal features. To address these challenges, we propose a Lightweight Adaptive Cue-Aware Vision Mamba network (LIDAR), which efficiently perceives and integrates morphological and textural cues from different modalities under multimodal crack scenarios, generating clear pixel-level crack segmentation maps. Specifically, LIDAR is composed of a Lightweight Adaptive Cue-Aware Visual State Space module (LacaVSS) and a Lightweight Dual Domain Dynamic Collaborative Fusion module (LD3CF). LacaVSS adaptively models crack cues through the proposed mask-guided Efficient Dynamic Guided Scanning Strategy (EDG-SS), while LD3CF leverages an Adaptive Frequency Domain Perceptron (AFDP) and a dual-pooling fusion strategy to effectively capture spatial and frequency-domain cues across modalities. Moreover, we design a Lightweight Dynamically Modulated Multi-Kernel convolution (LDMK) to perceive complex morphological structures with minimal computational overhead, replacing most convolutional operations in LIDAR. Experiments on three datasets demonstrate that our method outperforms other state-of-the-art (SOTA) methods. On the light-field depth dataset, our method achieves 0.8204 in F1 and 0.8465 in mIoU with only 5.35M parameters. Code and datasets are available at \url{https://github.com/Karl1109/LIDAR-Mamba}.
\end{abstract}





\maketitle

\section{Introduction}

Cracks with diverse morphologies frequently appear on the surfaces of real-world materials such as asphalt, concrete, plastic runways, and masonry, primarily due to shear stress. Thus, regular and automated structural health monitoring is essential to prevent losses in daily production and life~\cite{chen2024mind, liao2022automatic, lang2024augmented, zhang2023ecsnet, xue2023transformer, 10679891}. Recently, deep learning-based methods have demonstrated strong performance in automatic crack image segmentation~\cite{jaziri2024designing, liu2021crackformer, xiang2023crack, wang2024dual}. However, most methods rely solely on single-modality RGB data, making them vulnerable to lighting variations and background noise~\cite{liu2022asphalt_1}. They fail to capture subsurface thermal anomalies in infrared images, model stress-induced polarization changes, or interpret spatial hierarchies in depth images \cite{wang2022occlusion}, resulting in degraded performance under complex visual conditions such as uneven illumination, cluttered backgrounds, and ambiguous crack boundaries~\cite{liu2022asphalt_2}.

\begin{figure}[htbp]
  \centering
\includegraphics[width=0.44\textwidth]{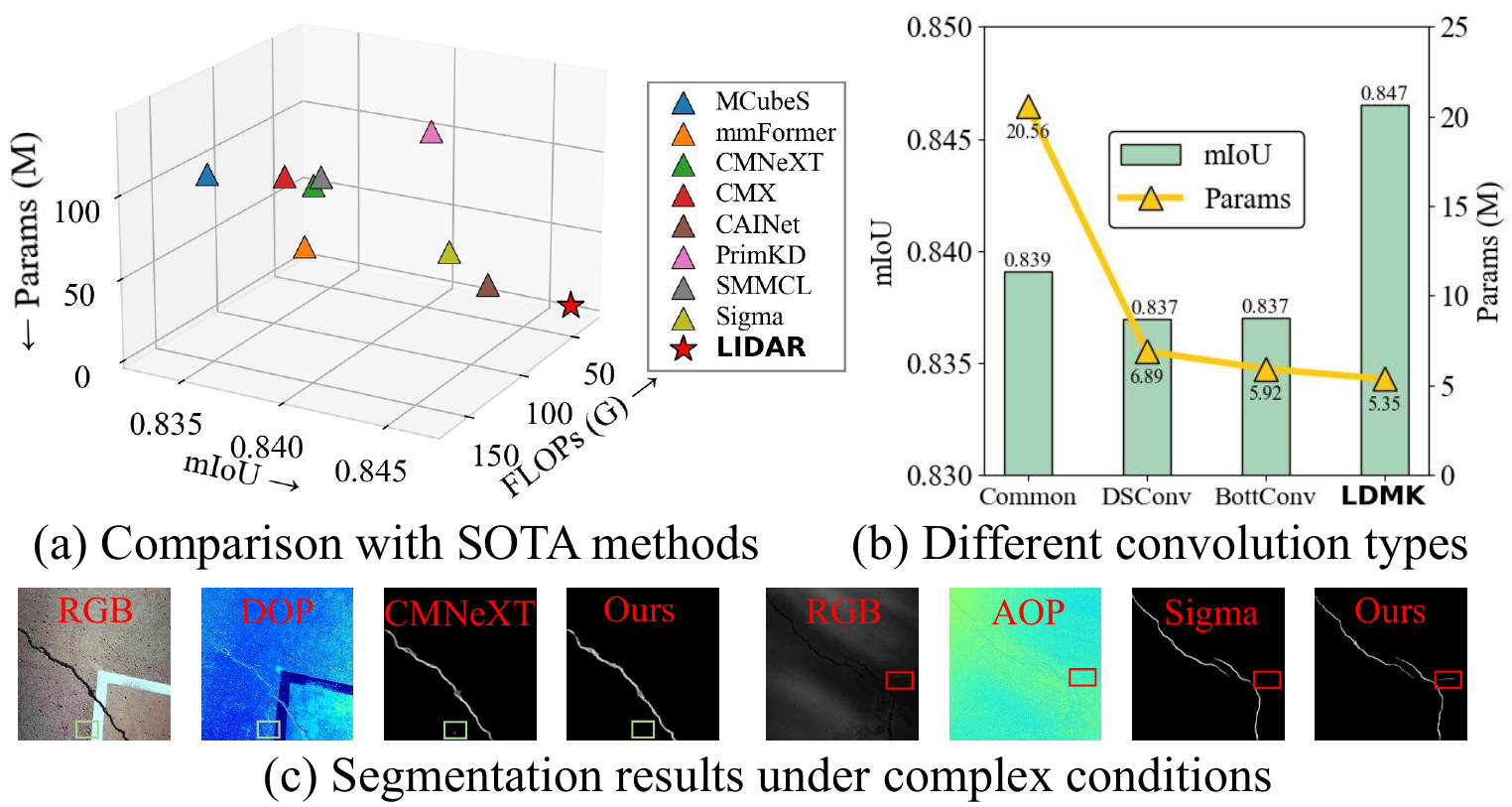}
  \caption{Performance of LIDAR on light-field depth dataset. (a) Comparison with SOTA methods. (b) Impact of different convolution types on performance. (c) Segmentation results for dual-modality images under complex conditions.}
  \label{fig:intro}
  \vspace{-0.6cm}
\end{figure}

Recent multimodal semantic segmentation methods based on Convolutional Neural Networks (CNNs) and Transformers have achieved promising results~\cite{yang2023pixel, ye2022taskprompter, bachmann2022multimae, wang2024pssd}. Representative methods such as PGDENet~\cite{zhou2022pgdenet}, CAINet~\cite{lv2024context}, and ESANet~\cite{seichter2022efficient} utilize CNNs to fuse high-level semantics with low-level spatial details through progressively guided strategies that reduce modality discrepancies. While CNN-based methods effectively capture morphological cues in key regions, their limited receptive fields and inductive biases hinder the modeling of continuous texture patterns. Additionally, lots of  convolution operations lead to increased computational overhead. Transformer-based methods such as Omnivore~\cite{girdhar2022omnivore}, EMMA~\cite{zhao2024equivariant}, and CMX~\cite{zhang2023cmx} encode different modalities into interactive embeddings and employ self-attention to model long-range dependencies. Although they capture both morphological and textural cues effectively, the quadratic scaling of attention mechanisms with input length makes training and inference on high-resolution images computationally expensive and unsuitable for edge deployment~\cite{yu2025mambaout}. Despite their promising results, all of the above methods lack selective interaction and noise suppression across modalities and feature levels, leading to the loss of critical details.

Recently, the Selective State Space model (S6) Mamba~\cite{gu2023mamba, gu2022train} has shown impressive performance in long-sequence modeling. Compared to traditional linear time-invariant state space models (S4)~\cite{gu2022parameterization, goel2022s}, Mamba offers greater flexibility for complex data while reducing computational cost~\cite{lu2025jamma, xiao2025spatialmamba, guo2024mambair, liu2024cmunet}. Its success in vision tasks, such as Vision Mamba~\cite{vim}, confirms its effectiveness in modeling both local cues and long-range dependencies by scanning fixed-size pixel patches within each Visual State Space (VSS) block. Vision Mamba has since inspired various Mamba-based variants~\cite{zhang20242dmamba, li2024alignmamba}, where the scanning strategy is key to capturing spatial dependencies across regions.  PlainMamba~\cite{yang2024plainmamba}, VMamba~\cite{liu2024vmamba}, and SCSegamba~\cite{liu2025scsegamba} adopt 2D parallel, bidirectional, or combined parallel-diagonal snake scanning to enhance the modeling of spatial continuity, multi-directional context, and complex textures. While such strategies improve irregular structure perception, they rely on uniform, rule-based scanning for all images, limiting adaptability to highly complex and image-specific texture cues and topologies. This often results in discontinuities or blurred segmentation outputs. Moreover, even with uniform rules, generating repeated scanning sequences for each image introduces latency and reduces efficiency. Sigma~\cite{wan2024sigma}, the first to introduce Mamba into multimodal segmentation, adopts an inefficient parallel scanning strategy without selective interaction or noise suppression during fusion, resulting in high computational cost and missed detections in high-frequency critical regions, as shown in Figure~\ref{fig:intro}(c).
Additionally, existing Mamba-based methods stack many VSS blocks, increasing parameter count and computational overhead, which can lead to redundancy and feature degradation. Common convolutions with heavy parameters are used in feature processing and final segmentation, limiting the potential for deployment on resource-constrained devices.

To address the above challenges, we propose the Lightweight Adaptive Cue-Aware Vision Mamba network (LIDAR), which efficiently captures morphological and textural crack cues across various modalities—including RGB images, infrared thermography, polarization informations, and light-field depth cues, while handling arbitrary input sizes with low computational cost. This enables the generation of high-quality crack segmentation maps across diverse scenarios. As shown in Figure \ref{fig:intro}(a), LIDAR achieves the best performance while requiring the minimal computational resources. To make Mamba adaptable to complex modality-specific crack features, we design the Lightweight Adaptive Cue-Aware Visual State Space module (LacaVSS). It integrates the Efficient Dynamic Guided Scanning Strategy (EDG-SS) based on pre-scanned masks, which significantly accelerates the generation of scanning sequences. EDG-SS dynamically prioritizes crack regions based on image content, improving both the efficiency of texture modeling and the accuracy of crack-background separation. To reduce the computational cost of convolutions, we introduce the Lightweight Dynamically Modulated Multi-Kernel convolution (LDMK), which employs a dynamic intermediate channel selection mechanism and an adaptive selective kernel strategy. This design captures morphological cues through multiple receptive fields while maintaining low complexity. As shown in the Figure \ref{fig:intro}(b), LDMK enables the LIDAR to achieve the best performance while maintaining a low number of parameters. To enable effective cross-modal and hierarchical feature fusion, we propose the Lightweight Dual Domain Dynamic Collaborative Fusion module (LD3CF). It incorporates an Adaptive Frequency Domain Perceptron (AFDP) to enhance high-frequency crack features and suppress low-frequency background noise in both horizontal and vertical directions. Together with a dual-pooling strategy and dynamic gating, LD3CF enables efficient multi-level interaction with low computational cost.

In summary, our main contributions are as follows:

\begin{adjustwidth}{2em}{0pt}
$\bullet$ We propose LIDAR for multimodal structural crack segmentation. Adaptively captures morphological and textural cues across multiple modalities at a low computational cost, generating high-quality segmentation maps.
\end{adjustwidth}

\begin{adjustwidth}{2em}{0pt}
$\bullet$ We design the LacaVSS  based on the proposed EDG scanning strategy, which enables efficient and adaptive modeling of crack texture cues. The LDMK convolution significantly reduces computational cost while enhancing the perception of morphological information. The LD3CF module generates high-quality segmentation maps through efficient perception of frequency and spatial-domain cues, as well as a multi-level cross-modal interaction mechanism.
\end{adjustwidth}

\begin{adjustwidth}{2em}{0pt}
$\bullet$ We evaluate LIDAR on three datasets. Experimental results demonstrate that LIDAR outperforms existing SOTA methods while maintaining minimal computational requirements.
\end{adjustwidth}

\vspace{-0.35cm}
\section{Related Works}
\subsection{Crack Segmentation Methods}

RGB-based crack segmentation methods have achieved promising results~\cite{liu2021crackformer, liu2019deepcrack, liu2025scsegamba}. For instance, SFIAN~\cite{cheng2023selective} selectively fuses high-resolution texture and low-resolution semantic information at multiple scales to capture crack geometries. MGCrackNet~\cite{zhang2024robust} adopts a learnable parallel CNN-Transformer hybrid module to repeatedly fuse global and local features. CIRL~\cite{chen2024mind} introduces a clustering-inspired representation learning strategy that transforms supervised learning into an unsupervised clustering paradigm to extract more discriminative features. However, these methods rely solely on single-modality RGB data, making them unable to capture subsurface thermal anomalies, stress–polarization correlations, and spatial or geometric information. Consequently, their performance deteriorates under complex lighting and environmental conditions.

Although no dedicated methods have been specifically developed for multimodal crack segmentation, CNN and Transformer-based methods have achieved strong performance in general semantic segmentation tasks \cite{peng2024group, wang2024pssd, zhou2024bsbp}. For example, CAINet~\cite{lv2024context} improves accuracy through context-aware inference and detail aggregation, while PDCNet~\cite{yang2023pixel} utilizes pixel-difference convolution and cascaded large kernels for cross-modal feature fusion. However, CNNs suffer from limited receptive fields and strong inductive biases, making it difficult to model continuous texture patterns, while their dense convolutions result in high computational cost. Transformer-based methods such as CMNext~\cite{zhang2023delivering} and MCubeSNet~\cite{liang2022multimodal} fuse complementary information from arbitrary modalities and enhance segmentation via region-guided filter selection. Yet, their self-attention mechanisms introduce quadratic complexity with respect to sequence length, hindering deployment on resource-constrained devices. Moreover, both CNN and Transformer-based approaches lack mechanisms for selective semantic interaction and noise suppression across modalities and feature levels. Consequently, critical cues in fine-grained regions may be overwhelmed by redundant information. Dedicated multimodal crack segmentation methods are needed to effectively capture essential cues across modalities and enhance performance.

\begin{figure*}[htbp]
  \centering
  \includegraphics[width=0.90\textwidth]{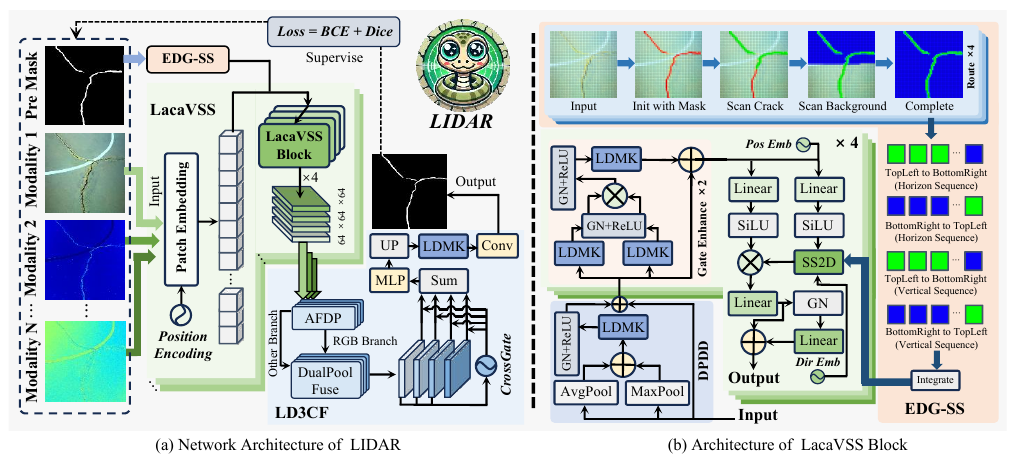}
  \caption{Overview of our LIDAR. Multimodal inputs are processed by LacaVSS to adaptively extract morphological and textural cues, and LD3CF fuses the multimodal feature maps to generate high-quality segmentation outputs. (a) illustrates the architecture of LIDAR and the processing flow for multimodal crack images. (b) illustrates the structure of LacaVSS.}
  \label{fig:LIDAR}
  \vspace{-0.4cm}
\end{figure*}

\vspace{-0.4cm}
\subsection{Selective State-Space Vision Model}
The Selective State Space model Mamba~\cite{gu2023mamba} has attracted attention for its strong performance in sequence modeling tasks. Compared to traditional linear time-invariant models (S4)~\cite{gu2022parameterization}, Mamba provides greater flexibility and computational efficiency for handling complex data, leading to its adoption in vision tasks. At its core, the VSS block performs block-wise scanning over feature maps to capture both fine-grained local details and long-range dependencies. The scanning strategy is crucial for capturing diverse structural and textural patterns. PlainMamba~\cite{yang2024plainmamba} uses direction-aware 2D parallel scanning to preserve semantic continuity, VMamba~\cite{liu2024vmamba} adopts bidirectional scanning to capture multi-directional dependencies, MaIR~\cite{MaIR} applies S-shaped scans within strip regions to maintain locality, and SCSegamba~\cite{liu2025scsegamba} combines parallel and diagonal snake scanning to enhance perception of complex textures. While these methods improve continuity perception through multi-path scanning, they rely on fixed scanning rules and lack the ability to adaptively generate scan sequences per input image. This limits their effectiveness in modeling highly variable textures, especially in crack segmentation where fine details are critical, often resulting in blurred or fragmented outputs. Moreover, repeated static path generation for each image introduces unnecessary latency, reducing inference efficiency. These networks also stack many VSS blocks and use high-parameter convolutions for feature extraction and segmentation, leading to substantial computational cost.

\vspace{-0.2cm}
\section{Method}
\subsection{Preliminary}

The overall architecture of our proposed LIDAR is illustrated in Figure \ref{fig:LIDAR}. It comprises two key components: the LacaVSS, which hierarchically extracts morphological and textural cues from different modal inputs, and the LD3CF, which captures frequency and spatial-domain information and generates high-quality segmentation maps through multi-level cross-modal interaction. Given $N$ input images from different modalities $\{X_1, X_2, ..., X_N\} \in \mathbb{R}^{B \times C \times 512 \times 512}$, where $B$ denotes the batch size and $C$ denotes the number of channels, each is first processed by a multi-layer LacaVSS backbone. The image is divided into $k$ patches, resulting in a sequence $\{P_1, P_2, ..., P_k\} \in \mathbb{R}^{B \times C  \times 8 \times 8 }$. These patches are scanned and processed by four LacaVSS blocks to extract morphological and textural crack features, producing feature maps $\{F_1, F_2, F_3, F_4\} \in \mathbb{R}^{B \times 64 \times 64 \times 64}$ for each modality. Finally, LD3CF fuses the feature maps across modalities and levels, generating the final segmentation  $\text{output} \in \mathbb{R}^{B \times 1 \times 512 \times 512}$.

\vspace{-0.3cm}
\subsection{Lightweight Dynamic Convolution}
\begin{figure}[htbp]
  \centering
  \includegraphics[width=0.4\textwidth]{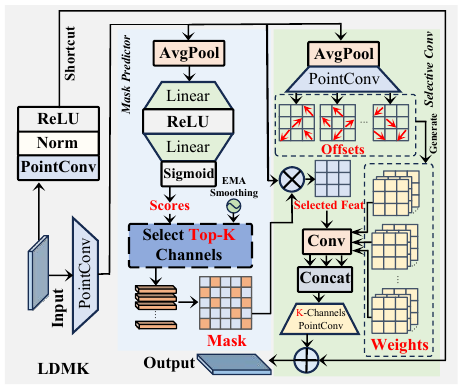}
  \caption{Architecture of LDMK. Adaptive multi-kernel feature extraction is applied to the Top-$K$ most important channels selected from the input features.}
  \label{fig:LDMK}
  \vspace{-0.4cm}
\end{figure}

The structure of LDMK is given by Figure \ref{fig:LDMK}. To reduce the parameter count and computational cost of convolution operations, LDMK adopts a channel modulation mechanism to dynamically select the most important feature channels for processing. This avoids redundant computation, significantly lowers resource consumption, and enables the extraction of critical crack-related morphological cues through multi-scale dynamic kernel selection across multiple receptive fields. Specifically, given an input feature map $\alpha \in \mathbb{R}^{B \times C_{\text{in}} \times H \times W}$, LDMK first applies a pointwise convolution to project the input from $C_{\text{in}}$ to an intermediate dimension $C_m$. It then models the importance of each channel. Concretely, the importance score $s$ for each channel can be obtained from the following equation:

\vspace{-0.5cm}
\begin{equation}
    \alpha_m = Conv_{1 \times 1}(\alpha) \in \mathbb{R}^{B \times C_m \times H \times W}
\end{equation}
\vspace{-0.3cm}
\begin{equation}
    s = \sigma \left( W_2 \cdot ReLU(W_1 \cdot AvgPool(\alpha_m)) \right) \in \mathbb{R}^{B \times C_m}
\end{equation}
where $\sigma(\cdot)$ denotes the Sigmoid activation function, and $W_1$, $W_2$ are learnable linear layer parameters. Then, the top-$k$ most influential channels are selected from $s$ to generate a binary mask $M \in \{0,1\}^{B \times C_m}$, which is used for channel-wise pruning:
\vspace{-0.1cm}
\begin{equation}
    \tilde{\alpha} = \alpha_m \odot M
\end{equation}
\vspace{-0.3cm}

To prevent instability caused by the sharp fluctuation in the number of active channels $k$ during training, LDMK adopts an Exponential Moving Average (EMA) strategy to smooth the channel activation ratio:
\begin{align}
    \hat{\rho}_t &= \gamma \cdot \hat{\rho}_{t-1} + (1 - \gamma) \cdot \rho_t, \quad \rho_t = \text{mean}(s)
\end{align}
where $\hat{\rho}_t$ denotes the smoothed activation ratio at iteration $t$, and $\gamma \in [0,1)$ is the EMA decay factor. During training, the number of activated channels $k_t = \lfloor C_m \cdot \hat{\rho}_t \rfloor$ is adjusted according to $\hat{\rho}_t$ to dynamically control the channel width.

We construct multiple shared depthwise convolution kernels $W_i \in \mathbb{R}^{C_m \times 1 \times k_i \times k_i}$, where kernel sizes $k_i \in \{3, 5, 7\}$ are used to capture texture features within different receptive fields. To enhance the adaptability of each receptive field, we introduce learnable scaling and shifting parameters $\alpha_i$ and $\beta_i$ for each branch:

\vspace{-0.3cm}
\begin{equation}
    \hat{W}_i = (1 + \alpha_i) \cdot W_i + \beta_i
\end{equation}
where $ \hat{W}_i $ denotes the dynamically reparameterized depthwise convolution kernel. Each branch incorporates the scaling factor $\alpha_i$ and bias term $\beta_i$ to enhance adaptiveness. The output features of each convolution branch are concatenated along the channel dimension, and the fused feature is subsequently processed by a pointwise convolution to recover the output dimension $C_{\text{out}}$, followed by addition with a residual connection.

\vspace{-0.2cm}
\subsection{Lightweight Adaptive Cue Aware}

The structure of LacaVSS is illustrated in Figure \ref{fig:LIDAR}(b). When all feature maps from a single modality are fed into the LacaVSS block, they are first processed by a Dual-Pooling Dynamic Denoiser (DPDD), which we designed to suppress local noise while preserving prominent structural information, thereby enhancing the stability and representational capacity of downstream modeling. Given the input feature map $\omega \in \mathbb{R}^{B \times C \times H \times W}$, the output of the average and max pooling operations is computed as:

\vspace{-0.3cm}
\begin{equation}
    \omega_1 = \text{AvgPool}(\omega) + \text{MaxPool}(\omega)
\end{equation}

Subsequently, $\omega_1$ is passed through the LDMK module to adaptively extract local information and produce the denoised output:

\vspace{-0.3cm}
\begin{equation}
\omega_{out} = \text{ReLU}(\text{GroupNorm}(\text{LDMK}(\omega_1))) + \omega
\end{equation}

To further enhance the extraction of morphological crack cues, a gating enhancement unit is integrated into the LDMK module. After two layers of processing, a sequence of patch features enriched with crack morphology is obtained.

Notably, before formal training, LIDAR utilizes pre-generated masks and the EDG-SS to produce a personalized adaptive scanning sequence for each set of multimodal crack images. During pre-training, the EDG-SS in LacaVSS is replaced by the base parallel scanning strategy, and the model is trained for 10 epochs on the multimodal crack dataset to obtain a pre-trained weight file. This weight file is then used to traverse all RGB images in the dataset and generate initial masks that represent general crack contours. These masks are subsequently used by EDG-SS to generate personalized scanning sequences for each multimodal image group before formal training. This process guides the model to better focus on the morphological and textural cues of crack regions.

The EDG-SS scanning procedure is shown in Figure \ref{fig:LIDAR}(b). Specifically, EDG-SS highlights salient regions based on the crack mask and incorporates an integral segmentation mechanism to rapidly assess the importance of each patch in various directions, thereby constructing adaptive scanning paths. This enables the model to prioritize regions with potential structural cues. Given a binary mask image $M \in \mathbb{R}^{H \times W}$ and patch size $p$, the integral image is defined as:

\vspace{-0.4cm}
\begin{equation}
I(x, y) = \sum_{i=0}^{x} \sum_{j=0}^{y} M(i, j)
\end{equation}

For a patch of size $p \times p$ starting from the top-left corner $(i, j)$, the importance score $ S_{i,j} $ can be computed as:

\vspace{-0.4cm}
\begin{equation}
S_{i,j} = I(i+p, j+p) - I(i+p, j) - I(i, j+p) + I(i, j)
\end{equation}

The EDG-SS separately traverses patches in horizontal and vertical directions to compute importance scores for each patch. Denote the importance score of the $k$-th patch in direction $d \in \{h, v\}$ as $S_k^{(d)}$, where $h$ and $v$ represent horizontal and vertical directions. Based on the importance values, all patches are divided into a crack region set $C_d$ and a background region set $B_d$, defined as:

\vspace{-0.3cm}
\begin{equation}
C_d = \{k \mid S_k^{(d)} > 0\}, \quad B_d = \{k \mid S_k^{(d)} = 0\}
\end{equation}

In each direction $d$, two scanning orders are defined: from the top-left to the bottom-right (tb) and from the bottom-left to the top-right (bt). By combining the two directions and two orders, EDG-SS constructs four scanning sequences. These four sequences are unified and defined as:

\vspace{-0.3cm}
\begin{equation}
\mathcal{O}_s^{(d)} =
\begin{cases}
[C_d] + [B_d], & \text{if } s = \text{tb} \\
[C_d]^{\text{rev}} + [B_d]^{\text{rev}}, & \text{if } s = \text{bt}
\end{cases}, \quad d \in \{h, v\}
\end{equation}
where $[\cdot]$ denotes the operation that converts a set into an ordered sequence of patch indices, $[\cdot]^{\text{rev}}$ denotes a reverse operation. EDG-SS ultimately generates four scanning sequences: $\mathcal{O}^{(h)}_{\text{tb}}$, $\mathcal{O}^{(h)}_{\text{bt}}$, $\mathcal{O}^{(v)}_{\text{tb}}$, and $\mathcal{O}^{(v)}_{\text{bt}}$. Unlike traditional scanning strategies, EDG-SS only needs to generate scanning sequences once during the preprocessing phase. All scanning sequences of the training and testing images are saved in a JSON file. During model inference or training, LacaVSS only needs to read the file once to retrieve the sequence corresponding to a specific image, avoiding redundant scanning and greatly saving computational time.

After generating all scanning sequences, the patch sequence is fed into the core SS2D module of LacaVSS. The input patch sequence is reordered according to the corresponding sequence $\mathcal{O}_s^{(d)}$, the position $\phi_{pos}$ and direction embedding $\psi_{dir}^{(d,s)}$ are added to each patch to form the input sequence:

\vspace{-0.2cm}
\begin{equation}
\eta^{(d,s)} = \eta	\left[ \mathcal{O}_s^{(d)} \right] + \psi_{\text{dir}}^{(d,s)} +
\phi_{\text{pos}}
\end{equation}
where $ \eta $ denotes the original input patch sequence extracted from the image. The sequence is subsequently passed into the core part of the LacaVSS block for state modeling, which is constructed based on discrete linear state space theory. At each scan position, the hidden state \( h_k \) is updated as follows:

\vspace{-0.3cm}
\begin{equation}
h_k = e^{\Delta A} h_{k-1} + \left[ \left( \Delta A \right)^{-1} \left( e^{\Delta A} - I \right) \cdot \Delta B \right] \eta_k
\end{equation}

\vspace{-0.2cm}
\begin{equation}
y_k = C h_k + D x_k
\end{equation}
where \( \eta_k \in \mathbb{R}^d \) denotes the input feature at the \( k \)-th scan position, \( h_k \) represents the current hidden state, and \( y_k \) is the model’s response at this position. \( \Delta A, \Delta B, C, D \) are learnable parameters used to dynamically adjust the update rate and strength of state propagation.

This modeling process is executed in parallel across all directions \( (d,s) \), resulting in four groups of state outputs, these outputs are aggregated and fused, and passed through a linear projection to obtain the output.

\vspace{-0.2cm}
\subsection{Lightweight Dual Domain Dynamic Fusion}

As illustrated in Figure \ref{fig:LIDAR}(a), the proposed LD3CF module aims to enhance and integrate multimodal features across levels by leveraging both frequency-domain perception and spatial-domain fusion.

The process begins with the AFDP, which receives the multimodal features extracted from LacaVSS. Given an input feature map $\zeta \in \mathbb{R}^{B \times C \times H \times W}$, a real-valued Fast Fourier Transform (rFFT) is first applied to project the features into the frequency domain. Direction-aware convolutions are then performed along horizontal and vertical axes to capture orientation-specific responses. 

To isolate discriminative frequency components, we introduce a learnable soft masking mechanism. Let $d_{\text{center}}^{h}$ and $d_{\text{center}}^{v}$ represent the distance from each frequency bin to the spectral center in horizontal and vertical directions, respectively. The corresponding high-frequency and low-frequency masks are defined as:

\vspace{-0.4cm}
\begin{equation}
\begin{split}
\mathcal{M}_{\text{high}}^{h,v} &= \sigma\left( (d_{\text{center}}^{h,v} - r) \cdot \tau \right), \\
\mathcal{M}_{\text{low}} &= 1 - \max\left( \mathcal{M}_{\text{high}}^h, \mathcal{M}_{\text{high}}^v \right)
\end{split}
\end{equation}
where $r$ is a learnable frequency separation radius, $\tau$ is a temperature scaling factor, and $\sigma(\cdot)$ denotes the Sigmoid function. These masks are used to reconstruct directional frequency components, which are then fused in the spatial domain through a channel-wise gating strategy, producing frequency-enhanced features.

Based on these refined features, LD3CF proceeds to perform dual-branch fusion to integrate information from multiple modalities. Let $\delta^{(0)}$ be the RGB modality feature and $\{\delta^{(l)}\}_{l=1}^{M-1}$ the auxiliary modality features, where M denotes the number of modals. The RGB mode is first enhanced by the following operations:
\begin{equation}
\delta_{RGB} = \delta^{(0)} \cdot \sigma\left( Linear \left( AvgPool(\delta^{(0)}) \right) \right)
\end{equation}

Then, each auxiliary modality interacts with $\delta_{\text{RGB}}$ through a dual pooling strategy:
\vspace{-0.1cm}
\begin{equation}
\delta_{fuse}^{(l)} = w_{1} \cdot AvgP(\delta_{RGB} + \delta^{(l)}) + w_{max} \cdot MaxP(\delta_{RGB} + \delta^{(l)})
\end{equation}
where $w_{1}$ and $w_{2}$ are learnable weights. The fused results are further transformed by LDMK convolutions to ensure compact and expressive representation. All modality-specific features are then aggregated to form a unified multimodal embedding:
\vspace{-0.1cm}
\begin{equation}
\delta_{sum} = \delta_{RGB} + \sum_{l=1}^{M-1} \delta_{fuse}^{(l)}
\end{equation}

To further ensure structural consistency and semantic complementarity across different feature levels, we introduce a cross-scale dynamic interaction mechanism. Let $\nu^{(n\in[0,3])}$ denote the output feature of LD3CF at level $n$, and $\nu^{(n-1)}$ the output from the previous level. These are adaptively fused via a learned gate:
\vspace{-0.1cm}
\begin{equation}
\nu^{(n)} = \nu^{(n)} \cdot G^{(n)} + \nu^{(n-1)} \cdot (1 - G^{(n)})
\end{equation}
\begin{equation}
G^{(n)} = \sigma \left(Linear \left(AvgPool\left(\nu^{(n-1)} \right) \right) \right)
\end{equation}
where $G^{(n)} \in [0,1]^{B \times C \times 1 \times 1}$ serves as a gate to balance semantic reinforcement and structural preservation. For the lowest level ($n=0$), no previous output exists, so $\nu^{(0)}$ is directly preserved without interaction.

All scale-level fused outputs are upsampled to the same spatial resolution and aggregated with learned weights. Subsequently, the result is processed by a linear layer, pointwise convolution, and further upsampling to generate the final segmentation map with rich awareness of crack shapes and texture structures.
\begin{table*}[htb]
\setlength{\tabcolsep}{1pt}
\caption{Performance comparison at dual-modal inputs. The best results are bolded and the second best results are underlined.}
\begin{tabular}{c|cccc|cccc|cccc|cccc}
\hline
\multirow{2}{*}{Method} & \multicolumn{4}{c|}{IRTCrack (RGB+Infrared)} & \multicolumn{4}{c|}{CrackDepth (RGB+Depth)} & \multicolumn{4}{c|}{CrackPolar (RGB+AOP)} & \multicolumn{4}{c}{CrackDepth (RGB+DOP)} \\ \cline{2-17} 
                        & ODS       & OIS       & F1        & mIoU     & ODS       & OIS       & F1       & mIoU     & ODS      & OIS      & F1       & mIoU     & ODS      & OIS      & F1       & mIoU    \\ \hline

CMX\cite{zhang2023cmx}                     & 0.8184    & 0.8244    & 0.8468    & 0.8463   & 0.8017    & 0.8032    & 0.8032   & 0.8336   & 0.7259   & 0.7376   & 0.7120   & 0.7847   & 0.7233   & 0.7261   & 0.7113   & 0.7843  \\
CAINet\cite{lv2024context}                 & 0.8216    & 0.8294    & 0.8367    & 0.8453   & 0.8054    & 0.8080    & 0.8044   & 0.8358   & 0.7327   & 0.7438   & 0.7174   & 0.7886   & 0.7326   & 0.7364   & 0.7180   & 0.7886  \\
PrimKD\cite{hao2024primkd}                  & 0.8271    & \underline{0.8345}    & 0.8457    & \underline{0.8518}   & 0.8010    & 0.8082    & 0.8013   & 0.8338   & 0.7385   & 0.7499   & 0.7258   & 0.7923   & 0.7378   & 0.7433   & 0.7279   & 0.7927  \\
SMMCL\cite{dong2024understanding}                   & 0.8090    & 0.8105    & 0.8388    & 0.8384   & 0.8019    & 0.8044    & 0.7996   & 0.8330   & 0.7289   & 0.7322   & 0.7210   & 0.7861   & 0.7284   & 0.7389   & 0.7213   & 0.7855  \\
SSRS\cite{ma2024multilevel}                   & 0.8198    & 0.8274    & 0.8420    & 0.8441   & 0.8143    & 0.8158    & 0.8133   & 0.8425   & 0.7314   & 0.7335   & 0.7157   & 0.7888   & 0.7264   & 0.7295   & 0.7118   & 0.7857  \\
Sigma\cite{wan2024sigma}                  & 0.8260    & 0.8322    & \underline{0.8545}    & 0.8513   & \underline{0.8168}    & 0.8176    & \underline{0.8147}   & \underline{0.8437}   & 0.7327   & 0.7456   & 0.7228   & 0.7895   & 0.7303   & 0.7320   & 0.7224   & 0.7875  \\
MCubeS\cite{liang2022multimodal}               & \underline{0.8284}    & \textbf{0.8350}    & 0.8430    & 0.8516   & 0.8167    & \underline{0.8217}    & 0.8114   & 0.8436   & \underline{0.7432}   & 0.7469   & 0.7299   & \underline{0.7960}   & \underline{0.7449}   & \underline{0.7522}   & \underline{0.7347}   & \underline{0.7972}  \\
CMNeXT\cite{zhang2023delivering}                  & 0.8256    & 0.8299    & 0.8406    & 0.8508   & 0.8031    & 0.8063    & 0.8026   & 0.8352   & 0.7359   & 0.7448   & 0.7256   & 0.7909   & 0.7405   & \textbf{0.7572}   & 0.7321   & 0.7927  \\
mmsFormer\cite{zhang2022mmformer}                & 0.8186    & 0.8297    & 0.8421    & 0.8448   & 0.8138    & 0.8180    & 0.8124   & 0.8419   & 0.7430   & \underline{0.7503}   & \underline{0.7307}   & 0.7951   & 0.7362   & 0.7437   & 0.7258   & 0.7913  \\
\rowcolor{gray!20}
\textbf{Ours}                    & \textbf{0.8305}    & 0.8316    & \textbf{0.8625}    & \textbf{0.8548}   & \textbf{0.8213}    & \textbf{0.8237}    & \textbf{0.8204}   & \textbf{0.8465}   & \textbf{0.7479}   & \textbf{0.7512}   & \textbf{0.7346}   & \textbf{0.8000}   & \textbf{0.7500}   & 0.7512   & \textbf{0.7382}   & \textbf{0.8015}  \\ \hline
\end{tabular}

  \label{tab:double_visual}
  \vspace{-0.2cm}
\end{table*}

\begin{figure*}[htbp]
  \centering
  \includegraphics[width=0.85\textwidth]{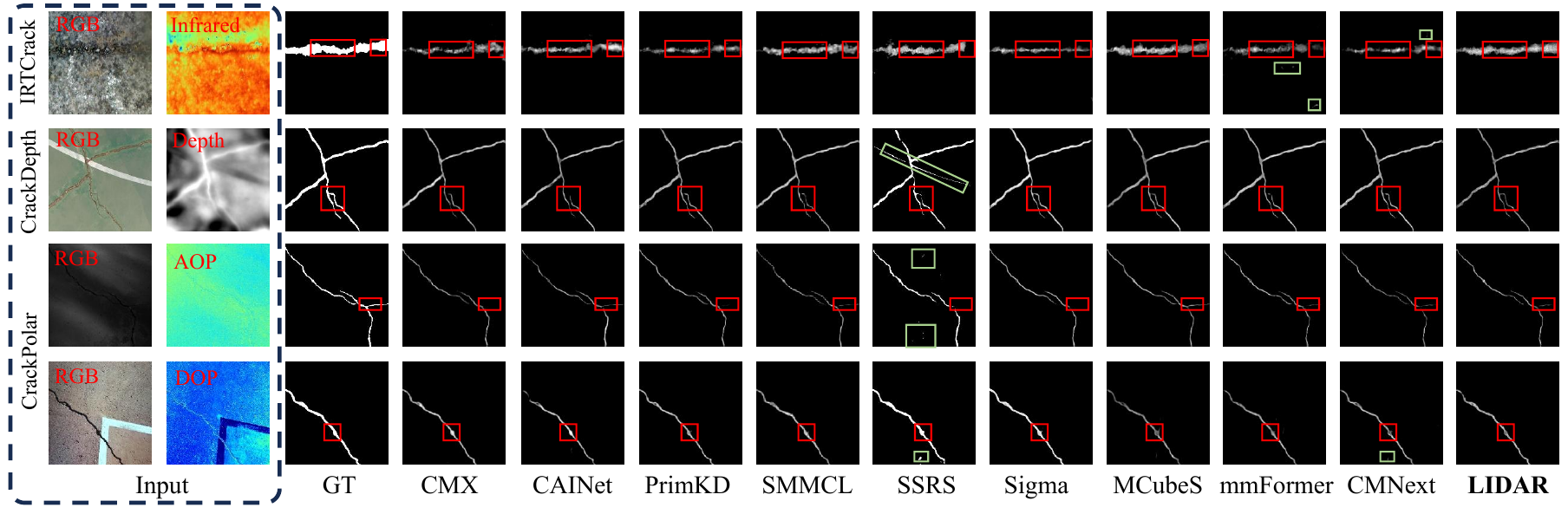}
  \caption{Visualization comparison for dual-modal inputs. Red boxes mark critical regions and green boxes mark noisy regions.}
  \label{fig:double_visual}
  \vspace{-0.4cm}
\end{figure*}

\vspace{-0.2cm}
\section{Experiments}
\subsection{Datesets}

\textbf{IRTCrack \cite{liu2022asphalt_1}}: The IRTCrack dataset consists of 448 paired RGB and infrared thermal images captured using a thermal imager. It covers diverse conditions, including various crack types, background textures, and lighting. Infrared thermography captures surface and subsurface thermal anomalies, enabling effective crack detection.

\noindent \textbf{CrackDepth}: CrackDepth is a dataset we collected, including 655 paired RGB and light-field depth images acquired via a light-field camera. Depth images are generated through post-processing. The dataset covers four surface types under different lighting, and the spatial and geometric cues in the depth data help distinguish crack morphology in complex scenarios.

\noindent \textbf{CrackPolar}: 
CrackPolar is a dataset we collected, containing 986 groups of images, including RGB images, Angle of Polarization (AoP) images, Degree of Polarization (DoP) images, as well as polarization images captured at four typical angles: 0°, 45°, 90°, and 135°, acquired using a polarization camera. It spans four material types under varying lighting. Polarized images highlight stress–polarization variations, enhancing crack-background contrast and detail visibility.

\vspace{-0.4cm}
\subsection{Implementation Details}

\textbf{Experimental Settings.} LIDAR is implemented using PyTorch v2.1.2 and trained on a server equipped with an Intel Xeon Platinum 8336C CPU and eight NVIDIA GeForce RTX 4090 GPUs running Ubuntu 20.04.6. During both pretraining and main training phases, we adopt the AdamW optimizer with an initial learning rate of 0.001. A polynomial learning rate decay strategy is used, and the weight decay is set to 0.01. LIDAR is pretrained for 10 epochs to generate initial structural masks, and then trained for 60 epochs in the main training phase. Input images are resized to a fixed resolution of 512$\times$512 before being fed into the network. The loss function uses the sum of the BCE \cite{jadon2020survey} and Dice \cite{sudre2017generalised} losses. All experiments were conducted under the same settings across all datasets.


\noindent \textbf{Evaluation Metrics.} We used four metrics to evaluate LIDAR's performance: F1 Score, Optimal Dataset Scale (ODS), Optimal Image Scale (OIS), and mean Intersection over Union (mIoU) \cite{liu2025scsegamba}. 

\vspace{-0.3cm}





\begin{table*}[htb]
\setlength{\tabcolsep}{1.8pt}
\caption{Performance comparison at multi-modal inputs. The best results are bolded and the second best results are underlined.}
\begin{tabular}{c|ccccccc|ccccccc}
\hline
\multirow{2}{*}{Method} & \multicolumn{7}{c|}{CrackPolar (RGB+{[}0°,45°,90°,135°{]})}        & \multicolumn{7}{c}{CrackPolar   (RGB+{[}0°,45°,90°,135°{]}+AOP)}   \\ \cline{2-15} 
                        & ODS    & OIS    & F1     & mIoU   & FLOPs   & Params  & Size   & ODS    & OIS    & F1     & mIoU   & FLOPs   & Params  & Size   \\ \hline
MCubeS\cite{liang2022multimodal}                  & \underline{0.7486} & \underline{0.7514} & \underline{0.7331} & \underline{0.7975} & 396.76G & 293.35M & 3594MB & 0.7483 & 0.7509 & 0.7323 & 0.7975 & 473.62G & 351.86M & 4045MB \\
mmsFormer\cite{zhang2022mmformer}                & 0.7403 & 0.7457 & 0.7275 & 0.7932 & \underline{76.39G}  & 59.99M  & 738MB  & 0.7356 & 0.7391 & 0.7214 & 0.7903 & \underline{84.94G}  & 73.34M  & 863MB  \\
CMNeXT\cite{zhang2023delivering}                  & 0.7473 & 0.7501 & 0.7302 & 0.7967 & \textbf{48.46G}  & \underline{57.66M}  & \underline{660MB}  & \underline{0.7515} & \textbf{0.7654} & \underline{0.7420} & \underline{0.7986} & \textbf{49.69G}  & \underline{57.67M}  & \underline{660MB}  \\
\rowcolor{gray!20}
\textbf{Ours}                    & \textbf{0.7503} & \textbf{0.7543} & \textbf{0.7383} & \textbf{0.8015} & 83.34G  & \textbf{13.35M}  & \textbf{193MB}  & \textbf{0.7576} & \underline{0.7647} & \textbf{0.7467} & \textbf{0.8023} & 100.01G & \textbf{16.01M}  & \textbf{231MB}  \\ \hline
\multirow{2}{*}{Method} & \multicolumn{7}{c|}{CrackPolar (RGB+{[}0°,45°,90°,135°{]}+DOP)}    & \multicolumn{7}{c}{CrackPolar (RGB+{[}0°,45°,90°,135°{]}+AOP+DOP)} \\ \cline{2-15} 
                        & ODS    & OIS    & F1     & mIoU   & FLOPs   & Params  & Size   & ODS    & OIS    & F1     & mIoU   & FLOPs   & Params  & Size   \\ \hline
MCubeS\cite{liang2022multimodal}                  & 0.7530 & 0.7600 & 0.7417 & 0.8000 & 473.62G & 351.86M & 4045MB & 0.7514 & 0.7555 & 0.7380 & \underline{0.7993} & 550.48G & 410.37M & 4496MB \\
mmsFormer\cite{zhang2022mmformer}                & 0.7523 & 0.7586 & 0.7384 & 0.8000 & \underline{84.94G}  & 73.34M  & 863MB  & 0.7445 & 0.7485 & 0.7296 & 0.7956 & \underline{93.50G}  & 86.68M  & 1034MB \\
CMNeXT\cite{zhang2023delivering}                  & \underline{0.7583} & \textbf{0.7642} & \underline{0.7449} & \underline{0.8035} & \textbf{49.69G}  & \underline{57.67M}  & \underline{660MB}  & \underline{0.7516} & \underline{0.7583} & \underline{0.7412} & 0.7991 & \textbf{50.92G}  & \underline{57.69M}  & \underline{660MB}  \\
\rowcolor{gray!20}
\textbf{Ours}                    & \textbf{0.7608} & \underline{0.7633} & \textbf{0.7476} & \textbf{0.8044} & 100.01G & \textbf{16.01M}  & \textbf{231MB}  & \textbf{0.7588} & \textbf{0.7668} & \textbf{0.7479} & \textbf{0.8031} & 116.68G & \textbf{18.68M}  & \textbf{270MB}  \\ \hline
\end{tabular}

\label{tab:multi_results}
\vspace{-0.1cm}
\end{table*}

\begin{figure*}[htbp]
  \centering
  \includegraphics[width=0.9\textwidth]{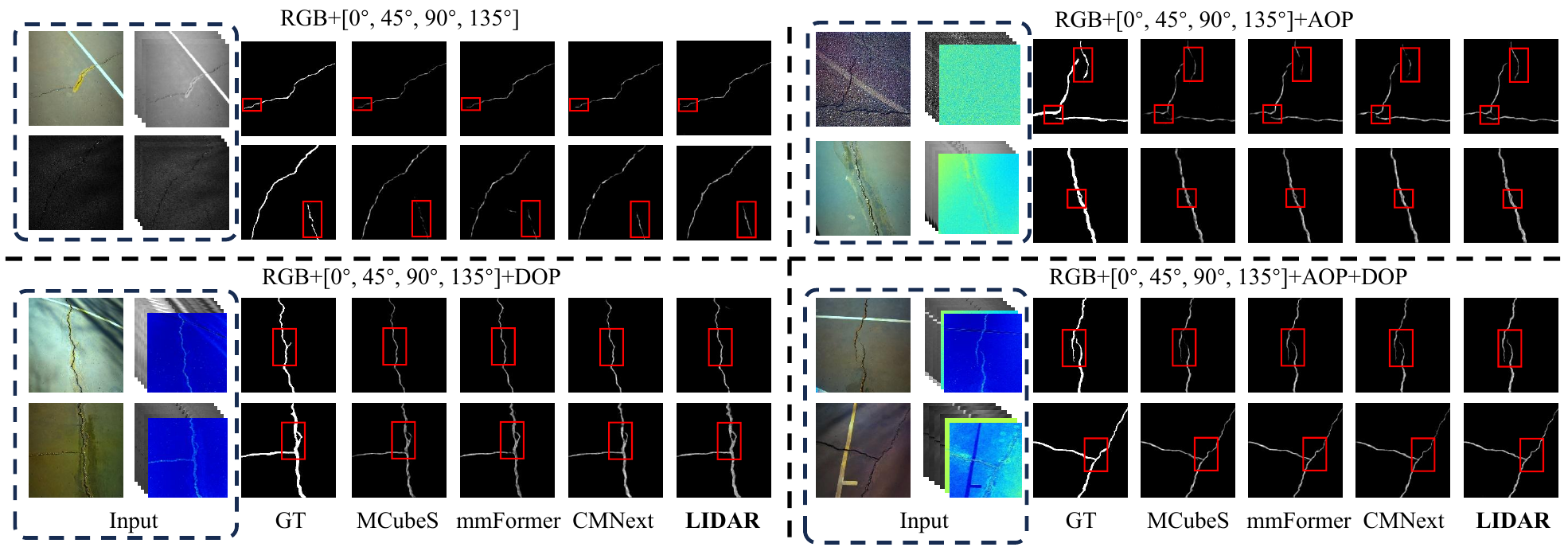}
  \caption{Visualization comparison for multi-modal inputs. Red boxes mark critical regions.}
  \label{fig:multi_visual}
  \vspace{-0.2cm}
\end{figure*}

\begin{table}[htb]
\setlength{\tabcolsep}{3pt}
\caption{Complexity comparison for dual-modal inputs.}
\begin{tabular}{c|c|ccc}
\hline
Method   & Year        & FLOPs   & Params  & Size   \\ \hline
MCubeS\cite{liang2022multimodal}   & CVPR 2022   & 165.59G & 117.76M & 2253MB \\
mmsFormer\cite{zhang2022mmformer} & MICCAI 2022 & 50.71G  & 19.96M  & 279MB  \\
CMNeXT\cite{zhang2023delivering} & CVPR 2023   & 44.76G  & 57.63M  & 660MB  \\
CMX\cite{zhang2023cmx}      & TITS 2023   & 56.99G  & 66.56M  & 762MB  \\
CAINet\cite{lv2024context}   & TMM 2024    & \underline{37.28G}  & \underline{10.31M}  & \underline{138MB}  \\
PrimKD\cite{hao2024primkd}   & MM 2024     & 114.42G & 139.85M & 1597MB \\
SMMCL\cite{dong2024understanding}    & WACV 2024   & 61.95G  & 72.86M  & 826MB  \\
SSRS\cite{ma2024multilevel}     & TGRS 2024   & 679.09G & 615.64M & 7322MB \\
Sigma\cite{wan2024sigma}    & WACV 2025   & 63.43G  & 41.68M  & 553MB  \\
\rowcolor{gray!20}
\textbf{Ours}     & MM 2025        & \textbf{33.33G}  & \textbf{5.35M}   & \textbf{78MB}   \\ \hline
\end{tabular}
\vspace{-0.3cm}
\label{tab:complex_dual}
\end{table}

\vspace{-0.3cm}
\subsection{Comparison with SOTA methods}

\noindent \textbf{Comparative experiments on two modality.}
As shown in Table~\ref{tab:double_visual} and Figure~\ref{fig:double_visual}, we evaluate LIDAR on IRTCrack \cite{liu2022asphalt_1},  CrackDepth and the CrackPolar using RGB, AoP, and DoP combinations. Our method consistently achieves superior performance across nearly all evaluation metrics. On the IRTCrack \cite{liu2022asphalt_1} dataset, LIDAR outperforms the second-best method by 0.94\% in F1 score and 0.35\% in mIoU, demonstrating its strong adaptability in modeling thermal crack responses. On CrackDepth, which includes light-field depth images, LIDAR surpasses Sigma \cite{wan2024sigma} by 0.55\%, 0.24\%, 0.70\%, and 0.33\% in ODS, OIS, F1, and mIoU, respectively, indicating its effectiveness in capturing spatial hierarchies and geometric structures. On the CrackPolar dataset, LIDAR maintains SOTA performance. With RGB+AoP or RGB+DoP input, it exceeds the second-best method by an average of 0.56\% in F1 and 0.52\% in mIoU. This highlights its capability to integrate polarization cues reflecting surface stress and microstructural patterns, enabling precise segmentation in fine-detail regions and robust background suppression.

As shown in Table \ref{tab:complex_dual}, LIDAR also achieves the lowest complexity under dual-modal input at 512$\times$512 resolution. Compared with CAINet \cite{lv2024context}, it reduces FLOPs, Params, and model size by 10.59\%, 48.10\%, and 43.48\%, respectively. These gains stem from the LDMK convolution, the LacaVSS, and the LD3CF module, which contributes only 0.25 GFLOPs and 0.02M Params. These components enable LIDAR to efficiently extract geometric and spatial cues across modalities and produce high-quality segmentation maps with detail continuity and noise suppression.

\noindent \textbf{Comparative experiments on multiple modalities.}
As shown in Table~\ref{tab:multi_results} and Figure~\ref{fig:multi_visual}, LIDAR achieves the best segmentation performance across all four polarization modality combinations compared to existing SOTA methods. When combining RGB with polarization angles (0$^\circ$, 45$^\circ$, 90$^\circ$, and 135$^\circ$), it surpasses the second-best method by an average of 0.71\% in F1 score and 0.50\% in mIoU. This highlights LIDAR’s ability to leverage reflectance differences across polarization angles for extracting critical morphological cues from diverse surface materials. When using RGB in combination with all four angles and either AoP or DoP, LIDAR further outperforms CMNext \cite{zhang2023delivering}, the second-best model, by 0.50\% in F1 and 0.29\% in mIoU on average. These results demonstrate LIDAR's strength in comprehensively capturing and fusing structural and textural cues encoded in polarization reflection behavior, microstructural directionality, and intensity variation.

\begin{table}[!t]
\vspace{-0.2cm}
\setlength{\tabcolsep}{1pt}
\caption{Comparison of different convolution types.}
\begin{tabular}{c|ccccccc}
\hline
Conv Type  & ODS    & OIS    & F1     & mIoU   & FLOPs   & Params & Size  \\ \hline
Common & \underline{0.8099} & 0.8126 & 0.8075 & \underline{0.8391} & 156.14G & 20.56M & 241MB \\
DSConv     & 0.8074 & 0.8115 & \underline{0.8082} & 0.8370 & 47.66G  & 6.89M  & 85MB  \\
BottConv   & 0.8076 & \underline{0.8137} & 0.8068 & 0.8370 & \underline{39.33G}  & \underline{5.92M}  & \textbf{74MB}  \\
\rowcolor{gray!20}
\textbf{LDMK} & \textbf{0.8213} & \textbf{0.8237} & \textbf{0.8204} & \textbf{0.8465} & \textbf{33.33G}  & \textbf{5.35M}  & \underline{76MB}  \\ \hline
\end{tabular}
\label{tab:conv_type}
\vspace{-0.5cm}
\end{table}

In terms of computational efficiency, while LIDAR incurs slightly higher FLOPs under multimodal input, it maintains the lowest parameter count and model size among all methods. For example, under full-modal input, LIDAR reduces parameter count and model size by 67.63\% and 59.09\% compared to CMNext \cite{zhang2023delivering}. This efficiency is due to the lightweight LDMK design, LacaVSS's adaptive modeling of crack topology and textures, and the LD3CF module’s capability to suppress irrelevant noise while enhancing key spatial and frequency-domain features. Together, these components enable LIDAR to produce high-quality segmentation maps with minimal computational overhead.

\begin{table}[!t]
\setlength{\tabcolsep}{2.5pt}
\caption{Performance comparison with scanning strategies.}
\begin{tabular}{c|ccccc}
\hline
Scan Type     & ODS    & OIS    & F1     & mIoU   & Delay Time \\ \hline
Para          & 0.7923 & 0.7935 & 0.7901 & 0.8269 & 1.63E-03s  \\
Diag          & 0.8013 & 0.8030 & 0.7984 & 0.8332 & 2.17E-03s  \\
ParaSnake       & 0.7993 & 0.8010 & 0.7971 & 0.8317 & 2.21E-03s  \\
DiagSnake       & 0.8067 & 0.8093 & 0.8033 & 0.8366 & 2.40E-03s  \\
bi\_ParaSnake   & 0.7976 & 0.8010 & 0.7946 & 0.8303 & 1.55E-03s  \\
bi\_DiagSnake   & 0.7989 & 0.8002 & 0.7980 & 0.8319 & \underline{6.25E-04s}  \\
SASS          & 0.8116 & 0.8158 & 0.8081 & 0.8406 & 2.16E-03s  \\
w/o pre       & \underline{0.8198} & \underline{0.8232} & \underline{0.8193} & \underline{0.8458} & 2.50E-02s  \\
w/o pre\&integral & 0.8189 & 0.8205 & 0.8186 & 0.8454 & 6.34E-02s  \\
\rowcolor{gray!20}
\textbf{EDG-SS}        & \textbf{0.8213} & \textbf{0.8237} & \textbf{0.8204} & \textbf{0.8465} & \textbf{7.15E-07s}  \\ \hline
\end{tabular}
\label{tab:ss_type}
\vspace{-0.4cm}
\end{table}

\vspace{-0.3cm}
\subsection{Ablation Studies}
We performed ablation experiments on the CrackDepth dataset.

\noindent \textbf{Performance comparison under different convolution types.} Table~\ref{tab:conv_type} lists the performance of LIDAR with different convolution types, including common convolution, DSConv \cite{sandler2018mobilenetv2} and BottConv \cite{liu2025scsegamba}. Our LDMK achieves the lowest FLOPs and parameter count while delivering the best performance across all metrics, including ODS, OIS, F1 score, and mIoU. Notably, compared to common convolution, LDMK reduces FLOPs, Params, and model size by 78.64\%, 73.97\%, and 68.05\%, respectively. These results confirm that LDMK not only enhances the extraction of morphological cues across modalities but also significantly lowers computational overhead.

\noindent \textbf{Ablation studies with different scanning strategies.} Table~\ref{tab:ss_type} lists the performance comparison of LIDAR using the proposed EDG-SS and several classical scanning strategies, including parallel, diagonal, parallel snake, diagonal snake, bidirectional scanning, and SASS \cite{liu2025scsegamba}. When EDG-SS is applied, LIDAR achieves the best results across all metrics, including ODS, OIS, F1 score, and mIoU, surpassing SASS by 1.19\%, 0.97\%, 1.52\%, and 0.70\%, respectively. It is noteworthy that models using conventional scanning strategies such as parallel, diagonal, and their snake variants suffer a clear performance drop. This indicates that fixed scanning paths lack adaptability and fail to effectively perceive the irregular texture cues present in different crack modalities, limiting the model’s capacity to characterize crack variations and suppress noise. While bidirectional scanning shows moderate improvement, the disruption of contextual dependencies in the central region of the image reduces its ability to model continuous semantic structures.

\begin{table}[!t]
\setlength{\tabcolsep}{2.5pt}
\caption{Performance comparison of different components combinations in LD3CF.}
\begin{tabular}{ccc|cccc}
\hline
AFDP & DualPool & CrossGate & ODS    & OIS    & F1     & mIoU   \\ \hline
\usym{2713}    & \usym{2717}        & \usym{2717}         & 0.7919 & 0.7932 & 0.7903 & 0.8265 \\
\usym{2717}    & \usym{2713}        & \usym{2717}         & 0.7894 & 0.7911 & 0.7891 & 0.8245 \\
\usym{2717}    & \usym{2717}        & \usym{2713}         & 0.7918 & 0.7929 & 0.7921 & 0.8273 \\
\usym{2713}    & \usym{2713}        & \usym{2717}         & 0.8073 & 0.8107 & 0.8050 & 0.8372 \\
\usym{2713}    & \usym{2717}        & \usym{2713}         & 0.8075 & 0.8094 & 0.8052 & 0.8371 \\
\usym{2717}    & \usym{2713}        & \usym{2713}         & \underline{0.8119} & \underline{0.8145} & \underline{0.8102} & \underline{0.8403} \\
\rowcolor{gray!20}
\usym{2713}    & \usym{2713}        & \usym{2713}         & \textbf{0.8213} & \textbf{0.8237} & \textbf{0.8204} & \textbf{0.8465}
\\ \hline
\end{tabular}
\vspace{-0.3cm}
\label{tab:ld3cf_com}
\end{table}

In terms of sequence generation latency, EDG-SS significantly outperforms all other strategies. It is 874 times faster (7.15E-07s VS 6.25E-04s) than the next-fastest method, bidirectional diagonal snake scanning. Compared with variants that omit the integral image or pre-scanning mechanism, EDG-SS achieves substantial acceleration. These results demonstrate that EDG-SS, by leveraging an integral-image-based pre-scanning mechanism and mask-guided path selection, effectively reduces latency and improves the model’s ability to capture continuous crack textures and suppress irrelevant background, thereby enhancing LIDAR’s segmentation performance in irregular crack regions.

\noindent \textbf{Ablation Study on Components of LD3CF.} Table~\ref{tab:ld3cf_com} lists the performance of LIDAR under different configurations of the LD3CF module components. Since LD3CF introduces only 0.25 GFLOPs and 0.02M parameters, the computational cost of each component is negligible. LIDAR achieved the best performance in all evaluation metrics when including all three components. Compared with the variant without AFDP, the complete model improves ODS, OIS, F1, and mIoU by 1.16\%, 1.13\%, 1.26\%, and 0.74\%, respectively. This demonstrates that AFDP enhances crack region representation by reinforcing high-frequency features and suppressing low-frequency background noise, leading to clearer textures for subsequent processing. Furthermore, removing both the dual pooling fusion and cross-level gating results in a significant performance drop, with F1 and mIoU decreasing by 3.81\% and 2.42\%, respectively. These findings indicate that the dual pooling fusion module effectively captures structural and texture cues in the spatial domain, while the cross-level gating mechanism enables adaptive and hierarchical interaction of multimodal features. Together, these components contribute to the generation of high-quality pixel-level crack segmentation maps.

\vspace{-0.3cm}
\section{Conclusion}

In this paper, we propose LIDAR, a pioneering Lightweight Adaptive Cue-Aware Vision Mamba network for pixel-level multimodal structural crack segmentation. LIDAR integrates the LacaVSS and LD3CF modules and replaces most convolutional operations with the proposed LDMK convolution, enabling efficient extraction and fusion of geometric, morphological, and textural cues across multiple modalities at low computational cost. LacaVSS, guided by the EDG-SS, dynamically prioritizes crack regions for more effective modeling. LD3CF enhances segmentation quality by combining AFDP and a dual pooling fusion module, enabling robust spatial-frequency cue extraction and background suppression, ultimately producing high-quality segmentation maps. Extensive experiments on three multimodal crack datasets demonstrate that LIDAR consistently outperforms SOTA methods in both performance and efficiency. For example, LIDAR demonstrates optimal performance across a variety of datasets and combinations of multiple modalities, consistently achieving best results in diverse scenarios. In future work, we aim to further improve LIDAR’s adaptability to modality-specific heterogeneity and explore more efficient learnable scanning strategies to enhance generalization on diverse crack datasets.

\vspace{-0.3cm}
\section{Acknowledgement}
This work was supported by the National Natural Science Foundation of China (NSFC) under Grants 62272342, 62020106004, 62306212, and T2422015; the Tianjin Natural Science Foundation under Grants 23JCJQJC00070 and 24PTLYHZ00320; and the Marie Skłodowska-Curie Actions (MSCA) under Project No. 101111188.


\clearpage
\appendix
\title{Supplementary material}

\section{Loss Function}
In crack segmentation tasks, the crack regions often occupy only a small portion of the image, leading to a severe class imbalance. To address this issue, we adopt a hybrid loss function that integrates a weighted Dice loss~\cite{sudre2017generalised} with Binary Cross-Entropy (BCE) loss~\cite{jadon2020survey}. The Dice loss emphasizes the spatial overlap between predictions and actual annotations, effectively alleviating class imbalance, whereas BCE focuses on enhancing the precision at the pixel level. By combining these two losses, the model benefits from both accurate detection of fine crack structures and robust global classification. This contributes to better delineation of crack edges and overall segmentation quality. The proposed loss formulation is defined as:

\vspace{-0.3cm}
\begin{equation}
Dice(p, \hat{p}) = 1 - \frac{2 \sum_{j=1}^{M} p_j \hat{p}_j + \epsilon}{\sum_{j=1}^{M} p_j + \sum_{j=1}^{M} \hat{p}_j + \epsilon}
\end{equation}

\begin{equation}
BCE(p, \hat{p}) = -\frac{1}{M} \sum_{j=1}^{M} \left[ p_j \cdot \log(\hat{p}_j) + (1 - p_j) \cdot \log(1 - \hat{p}_j) \right]
\end{equation}

\begin{equation}
Loss(p, \hat{p}) = Dice(p, \hat{p}) + BCE(p, \hat{p})
\end{equation}
where \( M \) denotes the total number of samples, \( p_j \) is the ground truth label for the \( j \)-th pixel, \( \hat{p}_j \) is the corresponding predicted probability, and \( \epsilon \) is a small constant added for numerical stability.

\section{Dataset Analysis}

\begin{figure}[!t]
  \centering
  \includegraphics[width=0.3\textwidth]{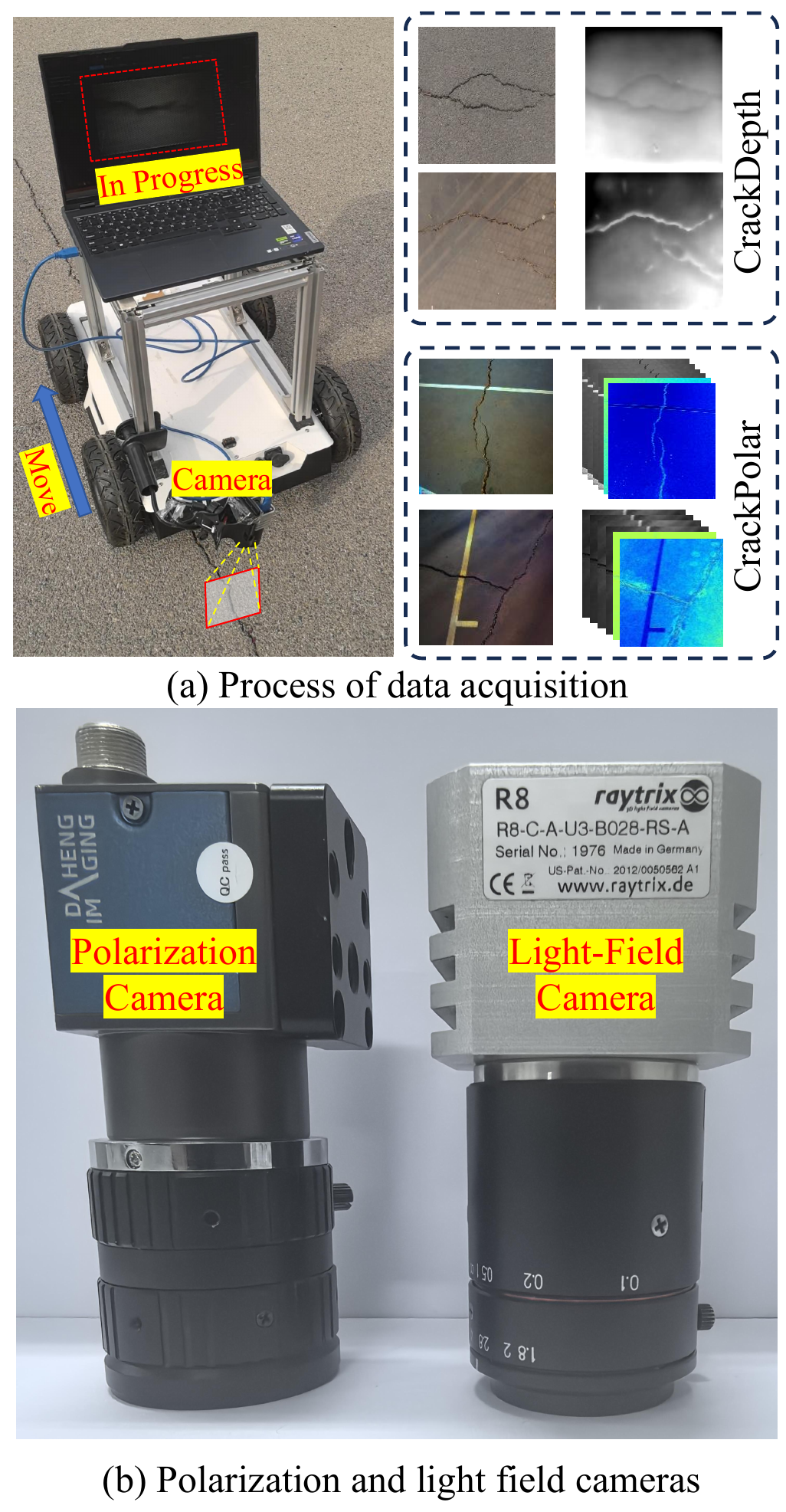}
  \caption{Demonstration of the dataset acquisition process.}
  \label{fig:Data_Acquisition}
  \vspace{-0.3cm}
\end{figure}

\begin{figure*}[htbp]
  \centering
  \includegraphics[width=0.9\textwidth]{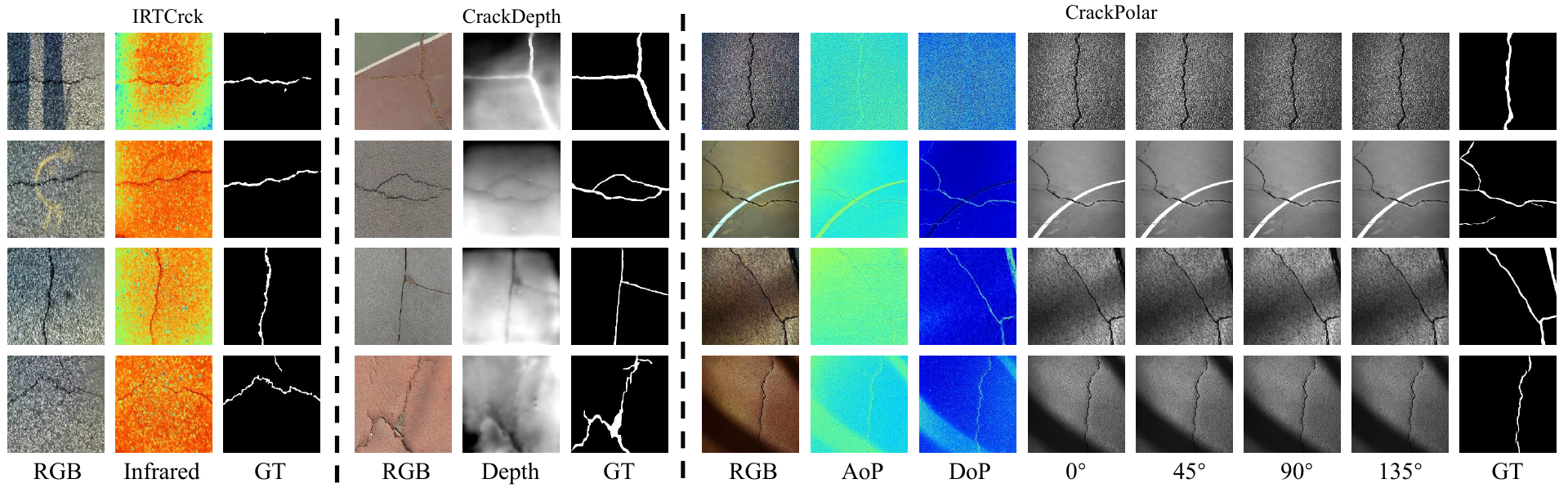}
  \caption{Visualization comparison of different datasets.}
  \label{fig:datasets_ana}
  \vspace{-0.2cm}
\end{figure*}

Existing multimodal crack datasets are extremely limited, with IRTCrack \cite{liu2022asphalt_1} being the only publicly available one. To address this gap and enable a more comprehensive evaluation of LIDAR, we construct two datasets: CrackDepth and CrackPolar. The acquisition process of the CrackDepth and CrackPolar datasets is shown in Figure~\ref{fig:Data_Acquisition}(a). Specifically, an intelligent vehicle was placed outdoors and remotely controlled to move at a constant speed while capturing crack images using onboard polarization and light-field cameras. The camera devices used are shown in Figure~\ref{fig:Data_Acquisition}(b). \textbf{The two datasets will be publicly available.} The comparison of these datasets is shown in Figure \ref{fig:datasets_ana} and Table \ref{tab:datasets_ana}. The following is the detailed analysis:

\textbf{IRTCrack} contains 448 pairs of RGB and infrared thermal images with a resolution of 640$\times$480, primarily collected from asphalt pavement environments. It includes diverse crack types, surface textures, and lighting conditions. The infrared thermal modality captures temperature anomalies on and beneath the surface, which is beneficial for identifying structural cracks induced by thermal diffusion or material degradation. Infrared imagery offers supplementary physical information beyond the visible spectrum, particularly effective under large temperature gradients or occlusions. In situations with uneven lighting or complex background textures, the thermal distribution patterns help stabilize responses in crack regions, aiding the detection of low-contrast or morphologically ambiguous cracks.

\begin{table}[htb]
\setlength{\tabcolsep}{0.4pt}
\caption{Comparison of different datasets.}
\begin{tabular}{c|c|c|c|c}
\hline
Dataset    & Modalities                                                                   & Num & Resolution & Scenarios                                                                     \\ \hline
IRTCrack   & RGB, Infrared                                                                & 448    & 640$\times$480    & Asphalt                                                                       \\ \hline
CrackDepth & RGB, Depth                                                                   & 655    & 512$\times$512    & \begin{tabular}[c]{@{}c@{}}Asphalt, Concrete, \\ Masonry, Runway\end{tabular} \\ \hline
CrackPolar & \begin{tabular}[c]{@{}c@{}}RGB, AoP, DoP, \\ 0°, 45°, 90°, 135°\end{tabular} & 986    & 512$\times$512    & \begin{tabular}[c]{@{}c@{}}Asphalt, Concrete, \\ Masonry, Runway\end{tabular} \\ \hline
\end{tabular}
\vspace{-0.3cm}
\label{tab:datasets_ana}
\end{table}

\textbf{CrackDepth} is an RGB-depth dataset containing 655 pairs of images at a resolution of 512$\times$512, collected using a light-field camera, with depth maps generated via post-processing \cite{wang2022occlusion}. It includes data from asphalt, concrete, masonry, and track surfaces under varying illumination conditions, enhancing generalization and difficulty. The depth modality inherently supports geometric modeling, allowing the capture of surface structure and elevation variations, which benefits the morphological characterization of cracks. Many real-world cracks manifest as shallow depressions or ridges not easily captured by RGB images. Depth maps provide discriminative supplementary information, especially in cases with shadows, complex textures, or structural occlusions, improving geometric awareness and segmentation accuracy.

\textbf{CrackPolar} is a large-scale multimodal crack image dataset built in this study, comprising 986 groups of images with a resolution of 512$\times$512. It includes RGB, Angle of Polarization (AoP), Degree of Polarization (DoP), and polarization images at four canonical angles: 0$^\circ$, 45$^\circ$, 90$^\circ$, and 135$^\circ$, captured using a polarization camera. The scenarios include asphalt, concrete, masonry, and track surfaces. The polarization modality captures light wave orientation and polarization states, effectively revealing stress-induced optical variations in material surfaces, thereby enhancing the contrast and contour clarity of crack regions. Compared to RGB imagery, polarization images are more sensitive to gloss, roughness, and stress concentrations, offering better robustness under weak textures, complex lighting, or high reflectivity. Additionally, AoP and DoP modalities enhance the modeling of directional consistency and structural coherence, supporting the extraction of more discriminative crack features through multi-scale and multi-angle information fusion.

\begin{figure}[!t]
  \centering
  \includegraphics[width=0.35\textwidth]{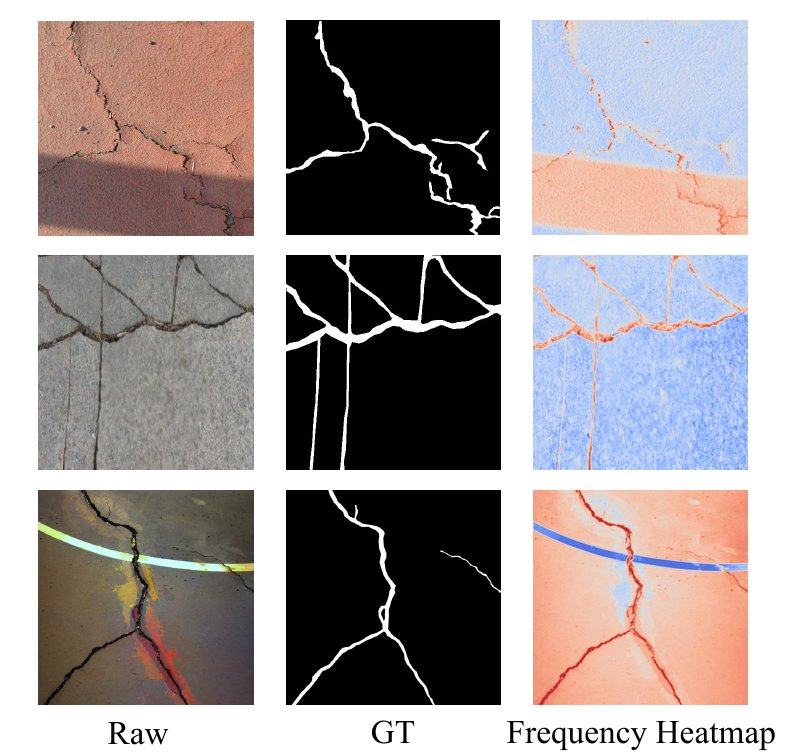}
  \caption{Comparison of RGB images of cracks, Ground Truth and frequency domain heatmaps. In the frequency domain heatmaps, darker red color indicates higher frequency and darker blue color indicates lower frequency.}
  \label{fig:Frequency_Heatmap}
  \vspace{-0.6cm}
\end{figure}

\begin{figure}[htbp]
  \centering
  \includegraphics[width=0.4\textwidth]{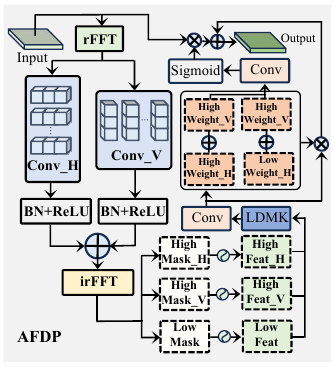}
  \caption{The structure of AFDP. AFDP enhances the high-frequency region of the crack and suppresses the low-frequency region of the background to provide downstream processing with a clearer feature map of the texture cues.}
  \label{fig:afdp}
    \vspace{-0.3cm}
\end{figure}

\section{Details of AFDP}

In structural crack images, crack regions typically exhibit slender shapes with well-defined edges. These local textural variations correspond to prominent high-frequency components in the frequency domain. In contrast, background regions often consist of flat or smoothly varying areas, primarily represented by low-frequency components in the frequency spectrum. Consequently, high-frequency signals tend to carry crucial information about crack boundaries and textures, while low-frequency components are mostly associated with irrelevant background or structural noise. As shown in Figure \ref{fig:Frequency_Heatmap}, the frequency heatmap visualization reveals stronger spectral responses in the crack regions, whereas the background exhibits weaker responses indicated by lighter tones. This observation demonstrates that crack features have high discriminative power in the frequency domain, offering reliable frequency cues to support effective segmentation.

To enhance the model's capability in capturing critical frequency-aware structures in crack images, we propose the AFDP. This module incorporates frequency domain transformation, directional soft masking, frequency-weighted filtering, and spatial fusion to selectively enhance high-frequency crack details while suppressing low-frequency background interference. The structure of AFDP is shown in Figure \ref{fig:afdp}.

Given an input feature map $x \in \mathbb{R}^{B \times C \times H \times W}$, we first apply a 2D real-valued Fast Fourier Transform (rFFT) to project it into the frequency domain. The frequency response reveals directional periodic patterns that are otherwise less distinguishable in the spatial domain, making it particularly effective for modeling textures and structural boundaries. The transformation is defined as:

\vspace{-0.3cm}
\begin{equation}
F(u, v) = \sum_{x=0}^{H-1} \sum_{y=0}^{W-1} f(x, y) \cdot e^{-j 2\pi \left( \frac{ux}{H} + \frac{vy}{W} \right)},
\end{equation}
where $f(x, y)$ denotes the pixel value in the spatial domain, $F(u, v)$ is its complex frequency representation, and $H$, $W$ are the height and width of the input. This transform decomposes the image into frequency bases, allowing the model to focus on discriminative frequency structures.

\begin{table*}[htb]
\setlength{\tabcolsep}{1.5pt}
\caption{Performance comparison at RGB unimodal input. The best results are bolded and the second best results are underlined.}
\begin{tabular}{c|cccc|cccc|cccc|ccc}
\hline
\multirow{2}{*}{Method} & \multicolumn{4}{c|}{IRTCrack (RGB)} & \multicolumn{4}{c|}{CrackDepth (RGB)} & \multicolumn{4}{c|}{CrackPolar (RGB)} & \multirow{2}{*}{FLOPs} & \multirow{2}{*}{Params} & \multirow{2}{*}{Size} \\ \cline{2-13}
                        & ODS     & OIS     & F1     & mIoU   & ODS     & OIS     & F1      & mIoU    & ODS     & OIS     & F1      & mIoU    &                        &                         &                       \\ \hline
CrackFormer\cite{liu2021crackformer}             & 0.7796  & 0.8001  & 0.8204 & 0.8185 & \underline{0.8111}  & \textbf{0.8234}  & \underline{0.8059}  & \underline{0.8393}  & 0.7038  & 0.7115  & 0.7104  & 0.7729  & 81.85G                 & 4.55M                   & 55MB                  \\
SimCrack\cite{jaziri2024designing}                & 0.7886  & 0.7977  & 0.8142 & 0.8242 & 0.7966  & 0.7981  & 0.7940  & 0.8291  & 0.7302  & 0.7377  & 0.7125  & 0.7889  & 286.62G                & 29.58M                  & 225MB                 \\
MambaIR\cite{guo2024mambair}                 & \underline{0.7962}  & 0.8036  & \underline{0.8297} & \underline{0.8294} & 0.7913  & 0.7937  & 0.7912  & 0.8254  & 0.7135  & 0.7188  & 0.7063  & 0.7777  & 47.32G                 & 10.34M                  & 79MB                  \\
CSMamba\cite{liu2024cmunet}                 & 0.7414  & 0.7456  & 0.7909 & 0.7925 & 0.7184  & 0.7203  & 0.7238  & 0.7753  & 0.6438  & 0.6478  & 0.6428  & 0.7342  & 145.84G                & 35.95M                  & 233MB                 \\
PlainMamba\cite{yang2024plainmamba}              & 0.7935  & 0.8000  & 0.8294 & 0.8277 & 0.7995  & 0.8029  & 0.8013  & 0.8306  & 0.7211  & 0.7292  & 0.7161  & 0.7824  & 73.36G                 & 16.72M                  & 96MB                  \\
SCSegamba\cite{liu2025scsegamba}               & 0.7961  & \underline{0.8084}  & 0.8214 & 0.8284 & 0.8032  & 0.8036  & 0.8008  & 0.8342  & \underline{0.7348}  & \underline{0.7396}  & \underline{0.7173}  & \underline{0.7925}  & \underline{18.16G}                 & \underline{2.80M}                   & \textbf{37MB}                  \\
\rowcolor{gray!20}
\textbf{Ours}                    & \textbf{0.8127}  & \textbf{0.8164}  & \textbf{0.8351} & \textbf{0.8408} & \textbf{0.8150}  & \underline{0.8166}  & \textbf{0.8139}  & \textbf{0.8420}  & \textbf{0.7362}  & \textbf{0.7406}  & \textbf{0.7294}  & \textbf{0.7929}  & \textbf{16.66G}                 & \textbf{2.68M}                   & \underline{39MB}                  \\ \hline
\end{tabular}

\label{tab:rgb_ana}
\end{table*}

\begin{figure*}[htbp]
  \centering
  \includegraphics[width=0.9\textwidth]{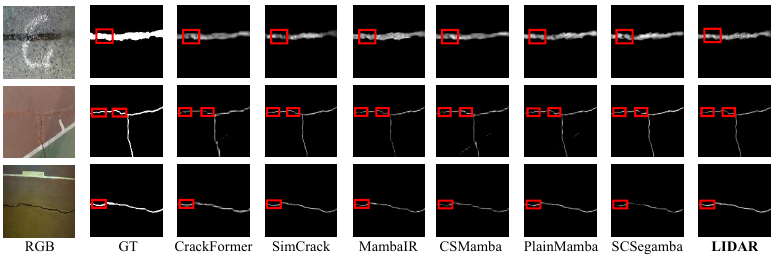}
  \caption{Visual comparison at RGB unimodal input. Red boxes mark critical regions.}
  \label{fig:rgb_visual}
  \vspace{-0.3cm}
\end{figure*}

AFDP then apply two convolutions along horizontal and vertical directions to independently model directional frequency components on both real and imaginary parts:

\vspace{-0.2cm}
\begin{align}
Freq_h &= ReLU(BN(Conv_h(FFT(x)))), \\
Freq_v &= ReLU(BN(Conv_v(FFT(x))))
\end{align}
where $Conv_h$ and $Conv_v$ are $1 \times 3$ and $3 \times 1$ convolutions, respectively. The outputs are merged and transformed back to the spatial domain via inverse rFFT.

To further improve frequency selectivity, we introduce a direction-aware soft masking mechanism based on the distance of each frequency bin to the spectral center. Let $d^h$ and $d^v$ represent the distances in horizontal and vertical directions. The high-frequency and low-frequency masks are defined as:

\vspace{-0.2cm}
\begin{equation}
\begin{split}
\mathcal{M}_{\text{high}}^{h} &= \sigma\left( \tau \cdot (d^{h} - r) \right), \\
\mathcal{M}_{\text{high}}^{v} &= \sigma\left( \tau \cdot (d^{v} - r) \right), \\
\mathcal{M}_{\text{low}} &= 1 - \max\left( \mathcal{M}_{\text{high}}^{h}, \mathcal{M}_{\text{high}}^{v} \right),
\end{split}
\end{equation}
where $\sigma(\cdot)$ is the Sigmoid function, $r$ is a learnable frequency separation radius, and $\tau$ is a temperature factor controlling the sharpness of the transition. This enables adaptive decomposition of high- and low-frequency components based on the input distribution.

After applying the masks, we obtain high-frequency components in both directions and the corresponding low-frequency background. Each is refined through lightweight LDMK convolutions to align channels. Next, direction-specific attention weights are generated using convolutions and Sigmoid activations to reflect the significance of each component. These weights are further fused into a unified channel-wise attention map that modulates the final response.

The output consists of a residual connection from the input modulated by the attention map, an enhanced high-frequency response guided by directional weights, and a suppressed low-frequency component. This composite output allows the model to retain global semantics while emphasizing fine-grained crack textures and boundaries, thereby improving segmentation performance in complex scenarios.

\section{Comparison with Single-Modal Methods}

In order to evaluate the robustness and generalizability of the proposed method without auxiliary modal information, we also compare LIDAR with unimodal SOTA methods under RGB unimodal input conditions, including CrackFormer \cite{liu2021crackformer}, SimCrack \cite{jaziri2024designing}, MambaIR \cite{guo2024mambair}, CSMamba \cite{liu2024cmunet}, PlainMamba \cite{yang2024plainmamba}, and SCSegamba \cite{liu2025scsegamba}.

As shown in Table~\ref{tab:rgb_ana} and Figure~\ref{fig:rgb_visual}, our proposed LIDAR consistently achieves the best performance across all three datasets compared to other single-modal SOTA methods. On the IRTCrack \cite{liu2022asphalt_1} dataset, LIDAR outperforms the second-best method by 0.65\% in F1 score and 1.38\% in mIoU, demonstrating the strong capability of LDMK in extracting morphological cues. On the CrackDepth and CrackPolar datasets, which contain finer cracks and more environmental interference, LIDAR still achieves the best segmentation performance, surpassing the second-best methods by 0.99\% and 1.69\% in F1 score, respectively. Compared with CSMamba \cite{liu2024cmunet} and SCSegamba \cite{liu2025scsegamba}, LIDAR produces more continuous predictions in fine-detail regions and better suppresses background noise. In terms of model complexity, under single RGB input, LIDAR requires only 16.66 GFLOPs and 2.68M parameters. This demonstrates that, with the efficient design of LDMK and EDG-SS, LIDAR can generate high-quality segmentation results with strong noise suppression and low computational cost.

\begin{table}[htb]
\setlength{\tabcolsep}{0.8pt}
\caption{Performance comparison of combining RGB, AoP and DoP modalities on CrackPolar. The best results are bolded and the second best results are underlined.}
\begin{tabular}{c|ccccccc}
\hline
\multirow{2}{*}{Method} & \multicolumn{7}{c}{CrackPolar (RGB+AOP+DOP)}                   \\ \cline{2-8} 
                        & ODS    & OIS    & F1     & mIoU   & FLOPs   & Params  & Size   \\ \hline
CMNeXT                  & 0.7425 & 0.7455 & \underline{0.7332} & 0.7946 & \textbf{45.99G}  & 57.64M  & 660MB  \\
MCubeS               & \underline{0.7435} & \underline{0.7458} & 0.7298 & \underline{0.7960} & 242.65G & 176.29M & 2703MB \\
mmFormer               & 0.7308 & 0.7386 & 0.7190 & 0.7868 & 59.28G  & \underline{33.31M}  & \underline{432MB}  \\
\rowcolor{gray!20}
\textbf{Ours}                    & \textbf{0.7442} & \textbf{0.7490} & \textbf{0.7416} & \textbf{0.7980} & \underline{50.00G}  & \textbf{8.02M}   & \textbf{116MB}  \\ \hline
\end{tabular}

\label{tab:crackpolar_rgb_aop_dop}
\end{table}

\begin{table*}[htb]
\caption{Performance comparison of different components
combinations in LacaVSS. The best results are bolded and the
second best results are underlined.}
\begin{tabular}{ccc|ccccccc}
\hline
DPDD & Dir\&Pos Embed & Gate Enhance & ODS    & OIS    & F1     & mIoU   & FLOPs  & Params & Size \\ \hline
\usym{2713}    & \usym{2717}        & \usym{2717}    & 0.8058 & 0.8099 & 0.8048 & 0.8365 & \underline{31.70G} & \underline{5.04M}  & \underline{65MB} \\
\usym{2717}    & \usym{2713}        & \usym{2717}    & 0.8058 & 0.8106 & 0.8029 & 0.8364 & \textbf{31.64G} & \textbf{5.02M}  & \textbf{64MB} \\
\usym{2717}    & \usym{2717}        & \usym{2713}    & 0.8042 & 0.8085 & 0.8008 & 0.8349 & 33.24G & 5.32M  & 75MB \\
\usym{2713}    & \usym{2713}        & \usym{2717}    & 0.8069 & 0.8116 & 0.8051 & 0.8372 & 31.71G & 5.05M  & \underline{65MB} \\
\usym{2713}    & \usym{2717}        & \usym{2713}    & 0.8104 & 0.8135 & 0.8074 & 0.8393 & 33.31G & 5.34M  & 76MB \\
\usym{2717}    & \usym{2713}        & \usym{2713}    & \underline{0.8119} & \underline{0.8156} & \underline{0.8103} & \underline{0.8400} & 33.26G & 5.32M  & 75MB \\
\rowcolor{gray!20}
\usym{2713}    & \usym{2713}        & \usym{2713}    & \textbf{0.8213} & \textbf{0.8237} & \textbf{0.8204} & \textbf{0.8465} & 33.33G & 5.35M  & 76MB \\ \hline
\end{tabular}

\label{tab:LacaVSS_Com}
\end{table*}

\begin{table}[htb]
\setlength{\tabcolsep}{2.5pt}
\caption{Performance comparison for different number of scan paths. The best results are bolded and the second best results are underlined.}
\begin{tabular}{c|c|ccccc}
\hline
Scan Num                       & N & ODS    & OIS    & F1     & mIoU   & Delay Time \\ \hline
\multirow{2}{*}{Para}          & 2 & 0.7907 & 0.7954 & 0.7891 & 0.8256 & 9.35E-04   \\
                               & 4 & 0.7923 & 0.7935 & 0.7901 & 0.8269 & 1.63E-03   \\ \hline
\multirow{2}{*}{Diag}          & 2 & 0.7982 & 0.8012 & 0.7961 & 0.8318 & 1.13E-03   \\
                               & 4 & 0.8013 & 0.8030 & 0.7984 & 0.8332 & 2.17E-03   \\ \hline
\multirow{2}{*}{ParaSnake}     & 2 & 0.7946 & 0.7976 & 0.7908 & 0.8282 & 1.02E-03   \\
                               & 4 & 0.7993 & 0.8010 & 0.7971 & 0.8317 & 2.21E-03   \\ \hline
\multirow{2}{*}{DiagSnake}     & 2 & 0.7995 & 0.8046 & 0.7983 & 0.8313 & 1.27E-03   \\
                               & 4 & 0.8067 & 0.8093 & 0.8033 & 0.8366 & 2.40E-03   \\ \hline
\multirow{2}{*}{bi\_ParaSnake} & 2 & 0.7865 & 0.7977 & 0.7840 & 0.8232 & 6.23E-04   \\
                               & 4 & 0.7976 & 0.8010 & 0.7946 & 0.8303 & 1.55E-03   \\ \hline
\multirow{2}{*}{bi\_DiagSnake} & 2 & 0.7953 & 0.7977 & 0.7928 & 0.8300 & \underline{5.98E-04}   \\
                               & 4 & 0.7989 & 0.8002 & 0.7980 & 0.8319 & 6.25E-04   \\ \hline
\multirow{2}{*}{SASS}          & 2 & 0.8084 & 0.8094 & 0.8055 & 0.8381 & 1.10E-03   \\
                               & 4 & 0.8116 & 0.8158 & 0.8081 & 0.8406 & 2.16E-03   \\ \hline
\multirow{2}{*}{\textbf{EDG(Ours)}} & \cellcolor{gray!20}2 & \cellcolor{gray!20}\underline{0.8162} & \cellcolor{gray!20}\underline{0.8202} & \cellcolor{gray!20}\underline{0.8127} & \cellcolor{gray!20}\underline{0.8431} & \cellcolor{gray!20}\textbf{7.15E-07}   \\
& \cellcolor{gray!20}4 & \cellcolor{gray!20}\textbf{0.8213}   & \cellcolor{gray!20}\textbf{0.8237}   & \cellcolor{gray!20}\textbf{0.8204}   & \cellcolor{gray!20}\textbf{0.8465}   & \cellcolor{gray!20}\textbf{7.15E-07}   \\ \hline
\end{tabular}

\label{tab:scan_num}
\end{table}

\section{Additional Experiments on CrackPolar}
To more comprehensively evaluate the performance of LIDAR on the multimodal dataset CrackPolar, we conducted experiments with a combination of RGB, AoP and DoP, a total of three modalities. The AoP modality, which provides the polarization light vibration direction information, helps accurately locate the directional structure of cracks in complex backgrounds. The DoP modality reflects the intensity variation of polarized light and enhances the contrast between crack edges and background regions. By combining these two polarization modalities with the RGB image, LIDAR is able to better capture the multi-scale and multi-angle features of cracks, thus significantly improving the model's robustness and crack segmentation accuracy in complex environments. As shown in the Table \ref{tab:crackpolar_rgb_aop_dop}, LIDAR outperforms all other methods in terms of ODS, OIS, F1, and mIoU, particularly with an F1 score 1.62\% higher than the second-best MCubeS \cite{liang2022multimodal}. Furthermore, compared to the other three SOTA multimodal segmentation methods, LIDAR achieved the lowest parameter count and model size with the three-modal input, which is 75.92\% and 73.15\% smaller than the second-lowest mmFormer \cite{zhang2022mmformer}, respectively. This demonstrates that our method can achieve excellent segmentation performance while using minimal computational resources to perceive the morphological and texture cues of cracks across different modalities.

\begin{figure}[htbp]
  \centering
  \includegraphics[width=0.46\textwidth]{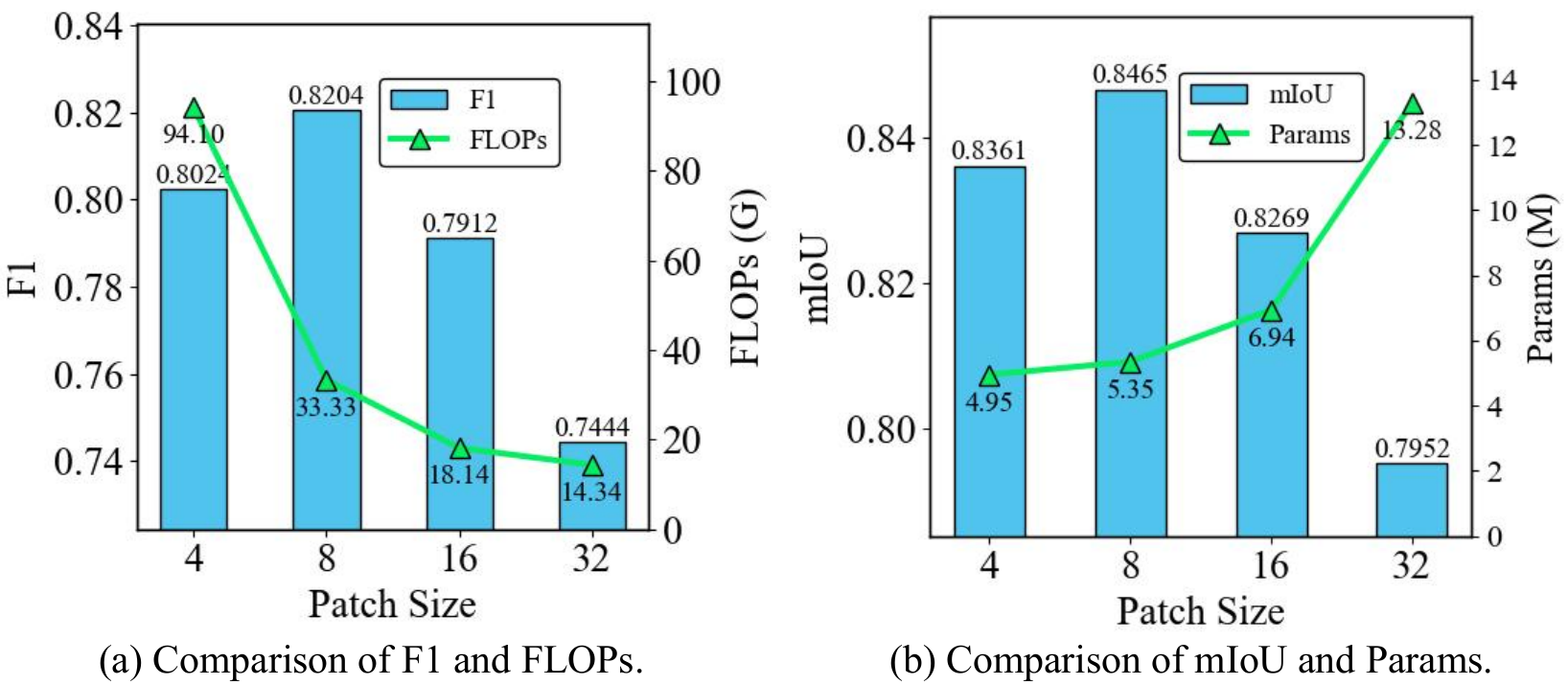}
  \caption{Comparison of different patch size.}
  \label{fig:patch_size_ana}
  \vspace{-0.3cm}
\end{figure}

\begin{table}[htb]
\setlength{\tabcolsep}{1.2pt}
\caption{Performance comparison for different patch sizes. The best results are bolded and the second best results are underlined.}
\begin{tabular}{c|ccccccc}
\hline
Patch Size & ODS    & OIS    & F1     & mIoU   & FLOPs  & Params & Size  \\ \hline
4          & \underline{0.8054} & \underline{0.8073} & \underline{0.8024} & \underline{0.8361} & 94.10G & \textbf{4.95M}  & \underline{85MB}  \\
\rowcolor{gray!20}
\textbf{8}          & \textbf{0.8213} & \textbf{0.8237} & \textbf{0.8204} & \textbf{0.8465} & 33.33G & \underline{5.35M}  & \textbf{76MB}  \\
16         & 0.7960 & 0.7960 & 0.7912 & 0.8269 & \underline{18.14G} & 6.94M  & 91MB  \\
32         & 0.7482 & 0.7503 & 0.7444 & 0.7952 & \textbf{14.34G} & 13.28M & 162MB \\ \hline
\end{tabular}

\label{tab:patch_size}
\end{table}

\section{Additional Experimental Analysis}
These experimental analyses were conducted on CrackDepth.

\noindent \textbf{Ablation Study on Components of LacaVSS.} 
The Table \ref{tab:LacaVSS_Com} lists the performance of LIDAR when different components of LacaVSS are used. When the DPDD, direction and position embedding, and gating enhancement mechanisms are fully utilized, LIDAR achieves the best performance. Compared to the variant without DPDD, the full model of LIDAR improves in ODS, OIS, F1, and mIoU, with FLOPs and Params only increasing by 0.07G and 0.03M, respectively. This indicates that our designed DPDD can suppress irrelevant local noise while preserving significant structure, thereby enhancing the modeling ability for downstream texture cues while maintaining lightness. Compared to the model without direction and position embedding, the full model improves F1 and mIoU by 1.25\% and 0.77\%, respectively, demonstrating that encoding the scan direction and position of the input Patch sequence allows the model to more accurately capture the crack arrangement direction and local position variations, thus improving the expression and discrimination of crack patterns. Notably, when only DPDD is used among the three components, the model requires the least computational resources, but its performance significantly decreases. Our designed DPDD, direction and position embedding, and gating enhancement mechanisms allow the model to effectively improve the modeling of crack morphology and texture cues in multimodal data, while maintaining low computational overhead.

\noindent \textbf{Analysis of Different Number of Scanning Paths.}
To evaluate the necessity of incorporating four directional scanning paths in the proposed EDG-SS, we conduct experiments by varying the number of scanning directions. Specifically the two paths of EDG-SS use horizontal scanning from top-left to bottom-right and vertical scanning from bottom-right to top-left respectively, while the other scanning strategies pick the default first two paths.

As shown in the Table \ref{tab:scan_num}, using four scanning paths consistently outperforms using only two paths across all strategies. For instance, when EDG-SS uses four paths, it surpasses the two-path variant by 0.95\% in F1-score and 0.40\% in mIoU. This performance gain is attributed to the richer crack details captured from multiple directions, enabling LacaVSS to integrate more comprehensive topological cues and enhancing the model’s ability to perceive complex crack morphologies. Notably, EDG-SS achieves the best performance regardless of whether two or four paths are used. When limited to two paths, EDG-SS outperforms the second-best SASS by 0.96\%, 1.33\%, 0.89\%, and 0.60\% in ODS, OIS, F1, and mIoU, respectively. Furthermore, EDG-SS demonstrates the lowest delay time in scanning path generation. In the two-path setting, it is 800× faster (7.15E-07 VS 5.98E-04) than the second-fastest dual-directional diagonal snake scan. This clearly demonstrates that EDG-SS, by leveraging an integral image-based pre-scanning strategy and a mask-guided mechanism, significantly reduces inference latency while more effectively extracting continuous texture cues from crack images, thereby enhancing LIDAR's capability in modeling irregular crack patterns and improving segmentation performance.

\begin{figure}[htbp]
  \centering
  \includegraphics[width=0.46\textwidth]{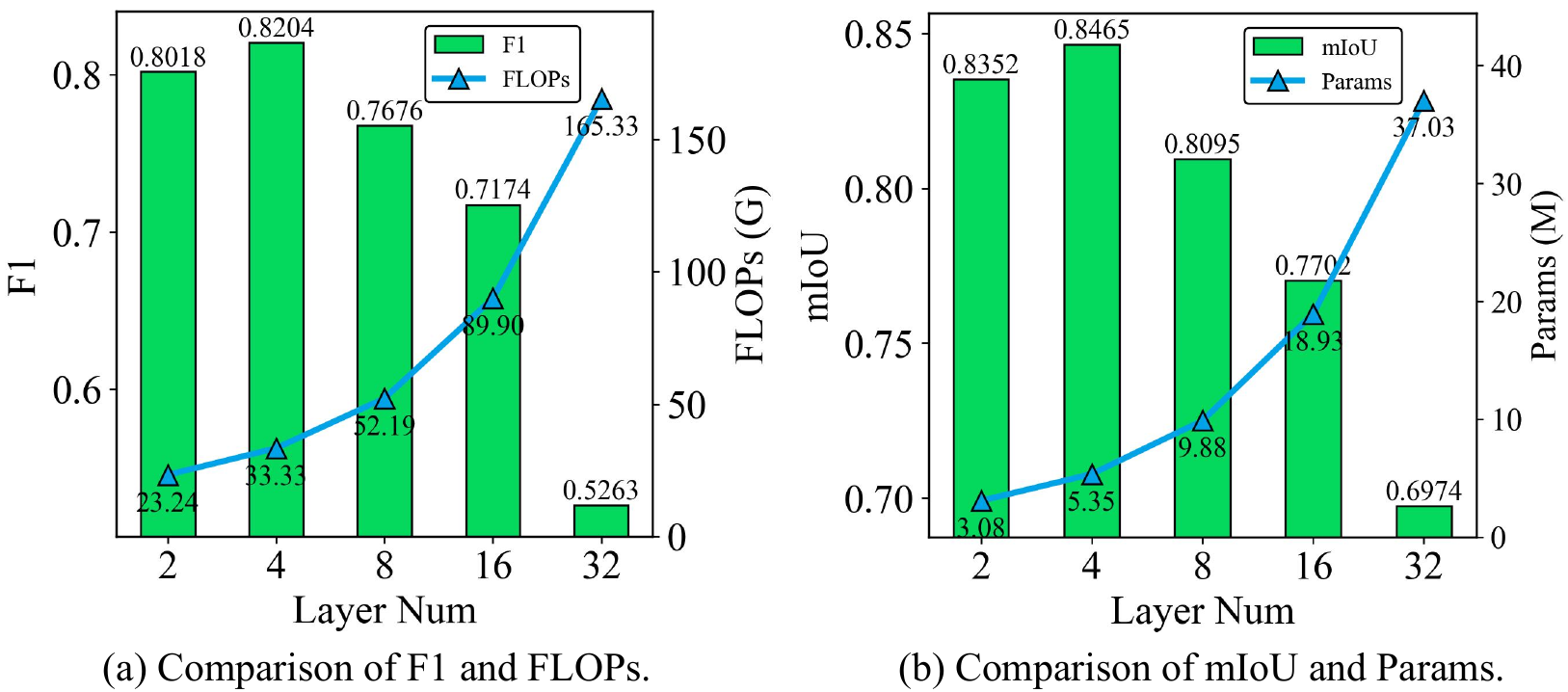}
  \caption{Comparison of different number of LacaVSS layers.}
  \label{fig:layer_num_ana}
  \vspace{-0.4cm}
\end{figure}

\begin{table}[htb]
\setlength{\tabcolsep}{1.2pt}
\caption{Performance comparison of different LacaVSS layers. The best results are bolded and the second best results are underlined.}
\begin{tabular}{c|ccccccc}
\hline
Layer Num & ODS    & OIS    & F1     & mIoU   & FLOPs   & Params & Size  \\ \hline
2         & \underline{0.8048} & \underline{0.8068} & \underline{0.8018} & \underline{0.8352} & \textbf{23.24G}  & \textbf{3.08M}  & \textbf{46MB}  \\
\rowcolor{gray!20}
\textbf{4}         & \textbf{0.8213} & \textbf{0.8237} & \textbf{0.8204} & \textbf{0.8465} & \underline{33.33G}  & \underline{5.35M}  & \underline{76MB}  \\
8         & 0.7561 & 0.7577 & 0.7676 & 0.8095 & 52.19G  & 9.88M  & 136MB \\
16        & 0.6896 & 0.7006 & 0.7174 & 0.7702 & 89.90G  & 18.93M & 257MB \\
32        & 0.5544 & 0.5602 & 0.5263 & 0.6974 & 165.33G & 37.03M & 498MB \\ \hline
\end{tabular}

\label{tab:layer_num}
\end{table}

\noindent \textbf{Analysis of Different Patch Sizes.}
When processing images from different modalities, LIDAR first divides each image into tokens using a sliding window with a patch size of 8×8 in the LacaVSS module. To validate the effectiveness of using a patch size of 8, we conducted experiments by varying the patch size. As shown in Table \ref{tab:patch_size} and Figure \ref{fig:patch_size_ana}, LIDAR achieved the best performance with a patch size of 8, outperforming the smaller patch size of 4 by 2.24\% in F1-score and 1.24\% in mIoU. This result indicates that smaller patch sizes constrain the local receptive field, making it difficult to capture widely distributed semantic features of irregular crack textures, thereby limiting performance.

Additionally, although increasing the patch size leads to a notable reduction in FLOPs, it also results in a significant increase in parameter count. This is because larger patch sizes yield fewer tokens, shortening the input sequence length, while each token covers a broader region, increasing the number of parameters needed for positional embedding. Notably, when the patch size was set to 32, LIDAR's performance dropped considerably. This decline is attributed to the limited sequence length, which prevents the LacaVSS from effectively capturing fine-grained crack morphological and textural cues via EDG-SS. Consequently, the model struggles to extract detailed semantic features, resulting in inferior segmentation performance. Therefore, setting the patch size to 8 provides a balanced trade-off between model efficiency and performance, allowing LIDAR to effectively perceive irregular geometric and textural features in multimodal crack images while maintaining lightweight computational costs.

\noindent \textbf{Analysis of Different LacaVSS Layers.}
We employed four layers of LacaVSS blocks in LIDAR to achieve a good trade-off between model efficiency and segmentation performance. To validate this design choice, we conducted experiments using different numbers of LacaVSS layers. As shown in Table \ref{tab:layer_num} and Figure \ref{fig:layer_num_ana}, LIDAR achieved the best performance with four LacaVSS layers, outperforming the variant with eight layers by 8.62\%, 8.71\%, 6.87\%, and 4.57\% in ODS, OIS, F1-score, and mIoU, respectively. Notably, although the model using only two LacaVSS layers required the least computational resources, with only 3.08M parameters, its performance dropped across all four metrics. As the number of LacaVSS layers increased, the FLOPs, parameter count, and model size grew significantly. In particular, using 32 layers resulted in not only extremely high computational overhead but also severely degraded performance, with the F1-score and mIoU dropping to just 0.5263 and 0.6974, respectively. This degradation is attributed to feature redundancy and overfitting in overly deep networks, which causes high-frequency crack features to be smoothed out during multi-level transmission, impairing the model's ability to effectively capture morphological and textural cues.

Furthermore, excessive LacaVSS stacking increases the difficulty of gradient propagation, leading to unstable training and performance deterioration. Therefore, selecting four LacaVSS layers provides an optimal depth that ensures sufficient feature modeling capacity while avoiding the redundancy and inefficiency associated with deeper networks. This choice enables LIDAR to achieve a favorable balance between accuracy and computational efficiency.

\bibliographystyle{ACM-Reference-Format}
\bibliography{sample-base}
\clearpage

\end{document}